\documentclass[10pt,twocolumn,letterpaper]{article}
\usepackage[pagenumbers]{cvpr}
\usepackage[dvipsnames]{xcolor}
\usepackage{multirow}
\usepackage[symbol]{footmisc}
\usepackage{tabularx}
\usepackage{caption}
\usepackage{adjustbox}
\usepackage{soul}
\usepackage{wrapfig}
\usepackage{epsfig}
\usepackage[accsupp]{axessibility}

\definecolor{cvprblue}{rgb}{0.21,0.49,0.74}
\usepackage[pagebackref,breaklinks,colorlinks,citecolor=cvprblue]{hyperref}

\def\paperID{21} 
\def\confName{CVPR}
\def\confYear{2024}

\title{Robustness Analysis on Foundational Segmentation Models}

\author{Madeline Chantry Schiappa\textsuperscript{1} \qquad  \qquad Shehreen Azad \textsuperscript{1*} \qquad  \qquad Sachidanand VS\textsuperscript{2} \\
Yunhao Ge\textsuperscript{3} \qquad \qquad Ondrej Miksik\textsuperscript{4} \qquad \qquad Yogesh S Rawat\textsuperscript{1} \qquad \qquad Vibhav Vineet\textsuperscript{4}
\\ 
\textsuperscript{1}Center for Research in Computer Vision, University of Central Florida\\\textsuperscript{2}Indian Institute of Technology, Madras;\hspace{.5cm}\textsuperscript{3}University of Southern California; \hspace{.5cm}\textsuperscript{4}Microsoft Research
}

\begin{document}
\twocolumn[{
\maketitle
\begin{center}
    \centering 
    % \vspace{-20pt}
    \captionsetup{type=figure}
    \includegraphics[width=\textwidth, height=6cm]{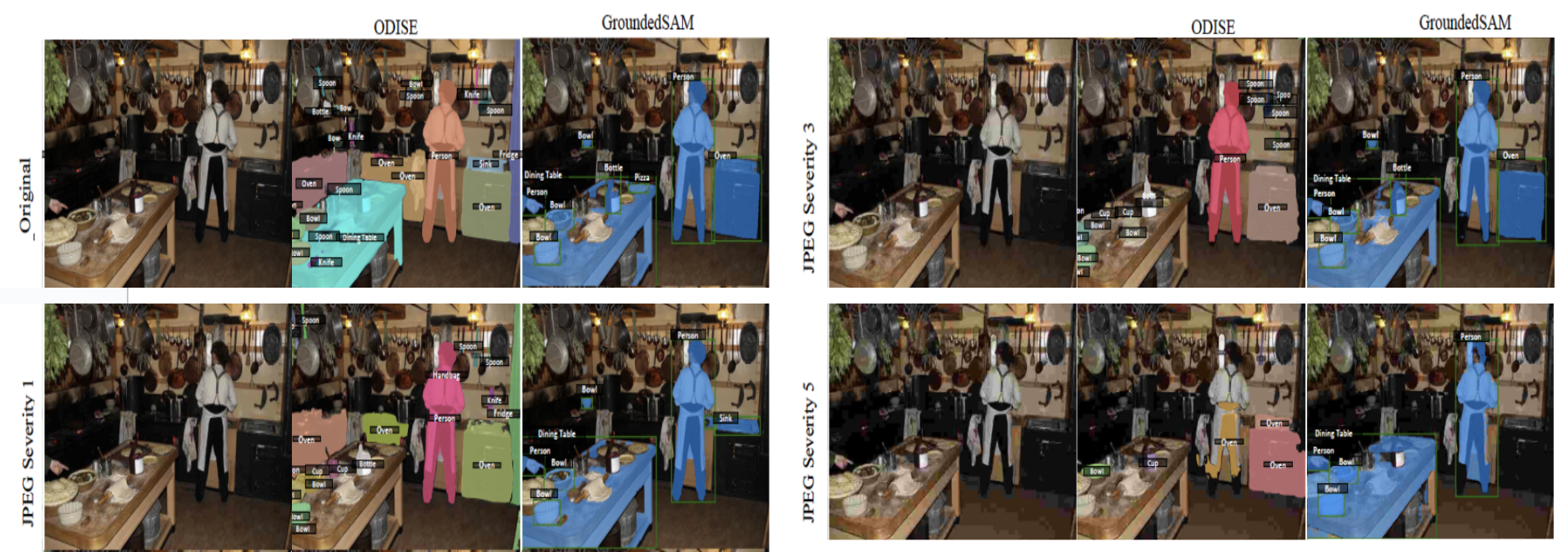}
    \caption{\textbf{Segmentation with foundation models} ODISE \protect\cite{odise} and GroundedSAM \protect\cite{ren2024grounded} on corrupted images (JPEG compression) at varying severity. An interesting observation is with corruption, while the person is still clearly visible, ODISE fails to recognize it even at severity 1.}
\label{teaser}
\end{center}
}] 

\begin{abstract}
Due to the increase in computational resources and accessibility of data, an increase in large, deep learning models trained on copious amounts of multi-modal data using self-supervised or semi-supervised learning have emerged. These ``foundation'' models are often adapted to a variety of downstream tasks like classification, object detection, and segmentation with little-to-no training on the target dataset. In this work, we perform a robustness analysis of Visual Foundation Models (VFMs) for segmentation tasks and focus on robustness against real-world distribution shift inspired perturbations. We benchmark seven state-of-the-art segmentation architectures using 2 different perturbed datasets, MS COCO-P and ADE20K-P, with 17 different perturbations with 5 severity levels each. Our findings reveal several key insights: (1) VFMs exhibit vulnerabilities to compression-induced corruptions, (2) despite not outpacing all of unimodal models in robustness, multimodal models show competitive resilience in zero-shot scenarios, and (3) VFMs demonstrate enhanced robustness for certain object categories. These observations suggest that our robustness evaluation framework sets new requirements for foundational models, encouraging further advancements to bolster their adaptability and performance. The code and dataset is available at: \url{https://tinyurl.com/fm-robust}. 
\footnotetext{*Corresponding Author: Shehreen.Azad@ucf.edu}

\end{abstract}

% %\vspace{-.5cm}
\section{Introduction}
\label{sec:intro}
Visual Segmentation is a longstanding challenge in computer vision, encompassing various tasks. These tasks require varying degrees of detail and include semantic \cite{long2015fully,chen2017deeplab, Zheng_2015_ICCV}, panoptic \cite{kirillov2019panoptic, zhang2023simple},  instance \cite{he2017mask, maskdino, wang2023internimage}. Traditionally, different tasks and datasets were handled independently with specialized models \citep{long2015fully, he2017mask, kirillov2019panoptic, qi2022open}, which did not allow cross-task synergy. However, with the advent of versatile transformer-based models \citep{vaswani2017attention, dosovitskiy2020image} and large-scale vision-language pre-training \citep{radford2021learning, chen2020simple, caron2021emerging, mae}, there's a growing shift towards developing comprehensive, multi-purpose, open-vocabulary vision systems, known as Visual Foundation Models (VFMs) \cite{yuan2021florence, li2022grounded, liu2023grounding}.

\noindent
In addition, inspired by the success of Large Language Models (LLMs), such as ChatGPT \cite{radford2021learning}, VFMs have harnessed the immense potential of foundation models and adapted them to open vocabulary instance segmentation tasks. For example, VFMs like Segment Anything (SAM) \cite{segmentanything}, ODISE \cite{odise} possesses the ability to segment any object within images without the requirement for further training. Such a breakthrough has ushered in many new opportunities in many safety critical real-world applications including autonomous vehicles, healthcare systems, etc. \cite{ma2023segment, yang2023track, ge2023building, deng2023segment}. 

\noindent
Deploying models in real world often introduces distribution shifts in the data, leading to unforeseen model behavior. To address this, studying the robustness of current deep learning models against potential real-world perturbations is essential \citep{comaniciu1997robust,kamann2020benchmarking,hendrycks2019benchmarking,schiappa2023large,ge2023improving, NEURIPS2022_de6ff07c}. 
These perturbations are not artificially induced through adversarial attacks, but naturally occurred due to changes in the environment, varying camera settings, and compression.
Hendrycks et.al. \cite{hendrycks2019benchmarking} introduced a series of such perturbations and evaluated the robustness of image classification models to these perturbations. Following this, such perturbations have been applied to evaluate robustness of models in several other downstream tasks \citep{kamann2021benchmarking,Schiappa_2023_CVPR}. 
While they studied the robustness of models in supervised settings for classification tasks; the robustness of VMFs for segmentation tasks remains uncertain regardless of the type of supervision during learning.  As VFMs become increasingly common and adapted for numerous downstream tasks, understanding their robustness and behavior in response to potential real-world distribution shifts is crucial.

\noindent
In this work, we conduct an extensive robustness analysis of VFMs with billions of parameters for the segmentation tasks. We use four recent multimodal VFM-based models, namely ODISE \cite{odise}, Painter \cite{Painter}, InternImage \cite{wang2023internimage}, Segment-Anything (SAM) \cite{segmentanything} along with recent unimodal models, namely Mask2Former \cite{mask2former}, MaskDINO \cite{maskdino} and ViT-Adapter \cite{vit-adapter}. For the first two non-VFM models, we use both CNN and transformer based backbones. For the robustness analysis, we use 17 common perturbations with 5 severity levels to the MS COCO \cite{coco} and ADE20K \cite{ade20k} datasets and name the perturbed datasets as MS COCO-P and ADE20K-P for the task of segmentation. 

\noindent
Our findings indicate that from the studied models, (1) VFMs lack robustness in compression and blur based corruptions. (2) All of the multimodal VFMs are not noticeably more robust nor higher performing in these segmentation tasks than the unimodal models; but have competitive robustness in a zero-shot setting. (3) multimodal VFMs show higher relative robustness for specific object-types compared to unimodal modelas. 

\noindent
In summary, our contributions are as follows:
\begin{itemize}
% %\vspace{-8pt}
\setlength\itemsep{0.01em}
    \item We focus on robustness analysis of foundational segmentation models against distribution shifts due to real-world inspired perturbations.
    \item We provide two benchmark datasets (MS COCO-P and ADE20K-P) to conduct robustness analysis on segmentation tasks. % We provide a benchmark dataset to study this problem. 
    \item We present an empirical analysis of foundational modeling approaches in segmentation to study the effect of various perturbations on their performance.
\end{itemize}

\section{Related Work}
\label{sec:rel_work}
\subsection{Vision Foundation Models}
The field of AI has seen a paradigm shift with the emergence of models trained on massive amounts of data at scale and are adaptable to various downstream tasks; which are commonly referred to as foundation models \cite{clip, imagebind, oquab2023dinov2, stablediffusion}. These models have shown remarkable performance in language and vision-related tasks, e.g. retrieval, recognition, segmentation etc.
Recent works on multi-modal learning have a trend of embedding features from different type of inputs to a common feature space \cite{clip, imagebind}; which is achieved by training these models using contrastive learning. Due to this common feature space of text in image embedding, it has been used for a huge number of downstream tasks \cite{stablediffusion,celebbasis,styleclip}. Stable diffusion \cite{stablediffusion} is one such popular model commonly used for generative purposes to solve the downstream tasks. To overcome the drawback of CLIP which overlooks the visual local information, DINOv2 \cite{oquab2023dinov2} is proposed, which is trained with self-supervised learning. A closed-set detector Dino \cite{dino} is extended into Grounding DINO \cite{groundingdino} for open-set object detection by performing vision-language modality fusion at multiple phases.

\noindent
Segmentation is an important computer vision task with safety-critical applications such as medical imaging and self-driving scenarios where robust models are required since wrong results could be catastrophic depending on the situation. There are some foundation models which have been developed for specific segmentation related tasks \cite{segmentanything, seem, wang2023internimage, Painter} some of which are based on the aforementioned models. However, while deployed in the real world the data these models come across might be corrupted with different kind of distribution shift, thus creating a necessity for robustness benchmarking for these models.

\subsection{Robustness}
Recently many work has focused on evaluating the vision model’s robustness in image \cite{hendrycks2019benchmarking,kamann2020benchmarking,bhojanapalli2021understanding,taori2020measuring,hendrycks2019augmix} and video domain \cite{schiappa2023large,Schiappa_2023_CVPR}. Hendrycks et al. \cite{hendrycks2019benchmarking} 
showed that the performance of bigger models gets affected just as the smaller models by corrupted data on the task of image classification. This shows that even though larger models may have more capacity to capture intricate features, they are not immune to the challenges posed by dataset variations or perturbations, leading to similar performance impacts as observed in smaller models. Following these works, such data corruptions were used for evaluating robustness of models in classification and object detection tasks. \cite{taori2020measuring,michaelis2019benchmarking}. 
The robustness of segmentation models has also been explored in recent works \cite{kamann2020benchmarking} using similarly perturbed dataset for evaluating the robustness of segmentation models highlighting a significant performance drop for corruptions affecting image texture versus those preserving it. 
There are similar work on video domain \cite{schiappa2023large} also evaluating the robustness of video action recognition models. There has also been work on improving model's robustness using augmentation techniques \cite{hendrycks2019augmix,geirhos2018imagenet,yin2019fourier,michaelis2019benchmarking}.

\noindent
With the growing popularity of the foundation model, performing very well in many areas of computer vision and with many downstream tasks being solved by using foundation model as an encoder, it is important to understand their behavior and robustness to potential real-world distribution shifts in the data. Towards this goal, we use a set of perturbations that are frequently encountered in real-world environments on datasets designed for segmentation tasks and evaluate the robustness of multi-modal models and compared it with unimodal models.

%\vspace{-.18cm}

\section{Experiments and Results}
\label{sec:robustness}
\subsection{Distribution Shifts and their Severity}
\label{perturbation_details}
Corruptions due to adversarial attacks are intentionally crafted to exploit vulnerabilities in machine learning models by adding imperceptible perturbations to the input data, thus causing misclassification or incorrect predictions. Unlike data corruption due to adversarial attack real-world data corruptions affect data during capture, transmission, or storage, and can degrade its quality. 

\noindent
In this work we study six different categories of real-world perturbations typically used in robustness benchmarking \cite{hendrycks2019benchmarking,NEURIPS2019_a2b15837,Schiappa_2023_CVPR,NEURIPS2022_de6ff07c,kamann2021benchmarking}. These categories include noise, blur, compression, digital, camera, and environmental perturbation and there is a total of \textit{17} different perturbations across all these categories. In noise, we have \textit{gaussian}, \textit{shot}, \textit{impulse}, and \textit{speckle} noise. In the blur category, we have \textit{defocus}, \textit{motion}, and \textit{zoom} blur. In the compression category, we have \textit{jpeg} and \textit{pixelate} corruption. In the digital category, we have \textit{contrast} and \textit{shear}. In the camera category, we have \textit{translate} and \textit{rotate} and finally, in the environment category, we have \textit{brightness}, \textit{darkness}, \textit{snow}, and \textit{fog}. The algorithms used to generate these corruptions follow previous literature \cite{hendrycks2019benchmarking, kamann2021benchmarking, NEURIPS2022_de6ff07c}.

\begin{figure*}
    \centering
    \includegraphics[width=\linewidth, height=5cm]{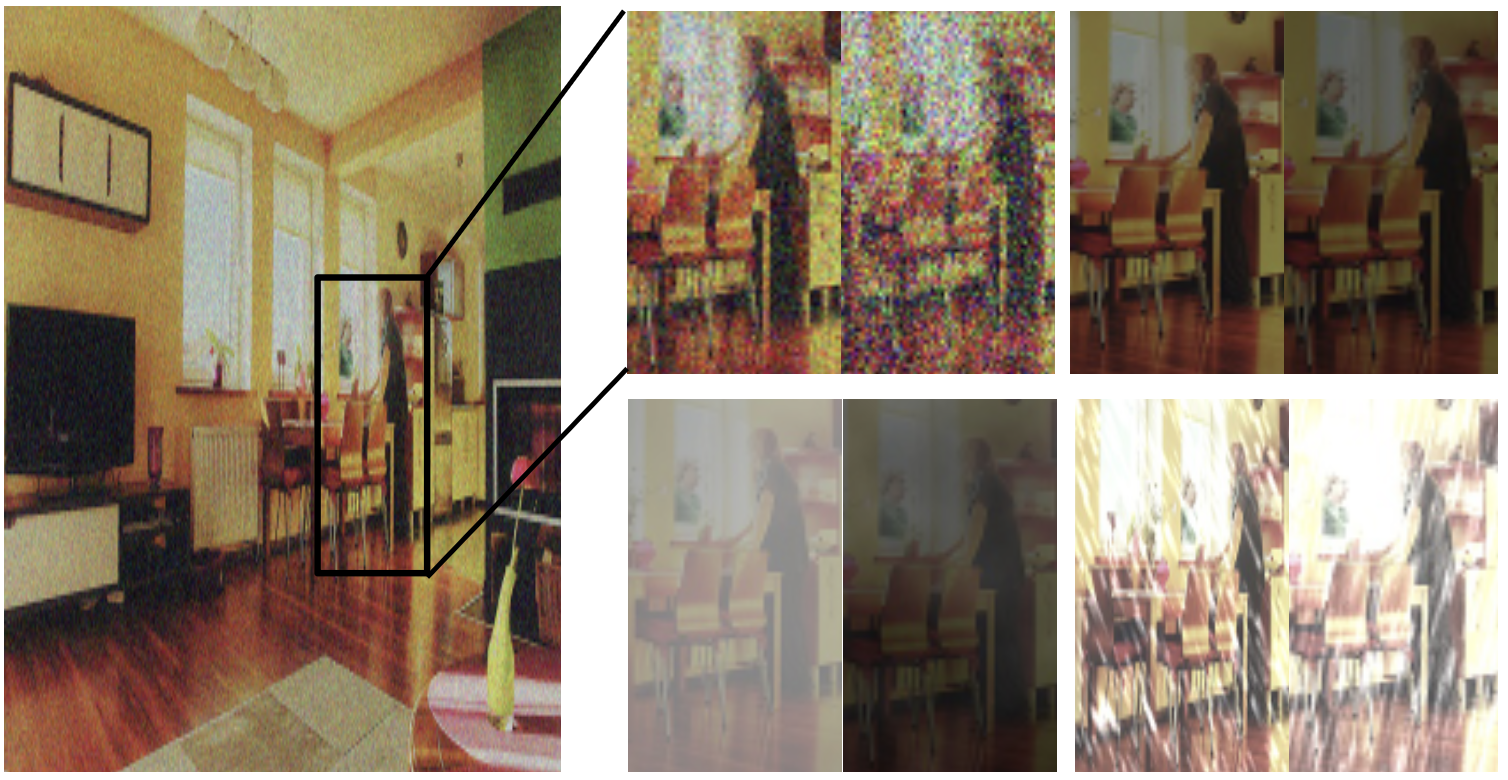}
    \caption{\textbf{Data perturbation examples} where original sample is zoomed in to show different corruptions on image from the MS COCO-P dataset. Each image pair is of corruption at severity 3 and 5. Top row shows corruptions in the category of gaussian noise and darkness, whereas, bottom row shows fog and snow.}
    \label{fig:example_Severity}
\end{figure*}

\noindent
In the real world, distribution shift corruptions may occur in varying levels of severity depending on the environment and/or situation. Therefore it is important to evaluate models under the same assumption that corruptions can vary in severity.  We generate five levels of severity where 1 is a small shift and 5 is a large distribution shift (Figure \ref{fig:example_Severity}). We apply all the proposed corruptions for each severity on all images using the \textit{imgaug} \cite{imgaug} library and code available from \cite{NEURIPS2022_de6ff07c} to generate the corruptions and their corresponding annotation. 

\subsection{Model Variants}
\label{model_details}
We perform our benchmark evaluation on seven state-of-the-art methods. We selected a set of models that were representative of multimodal Visual Foundation Models (VFMs) (ODISE \cite{odise}, two variations of Segment-Anything \cite{segmentanything}- PromptSAM \cite{promptsam} and GroundedSAM \cite{ren2024grounded}, InternImage \cite{wang2023internimage}, Painter \cite{Painter}) and comparative state-of-the-art unimodal models (ViT-Adapter \cite{vit-adapter}, Mask2Former \cite{mask2former} and MaskDINO \cite{maskdino})for segmentation. We selected models based on their availability of code, weights, and reproducability. 

\noindent
\textbf{ODISE} \cite{odise} is based on the feature space learned in Stable Diffusion \cite{stablediffusion} for their image encoder, CLIP \cite{clip} for their image-text discriminator and Mask2Former \cite{mask2former} as their mask generator. The process starts with extracting image features from Stable Diffusion with a \textit{Implicit Captioner} to learn implicit text prompts. These embeddings are passed to the mask generator. Similarity is measured between each mask and text embeddings of object categories from CLIP \cite{clip} to assign a class to a mask. While the model is trained on one dataset, it can be applied to any dataset for zero-shot evaluation, making it a strong model to consider. 

\noindent
\textbf{Segment-Anything} (SAM) \cite{segmentanything} uses a MAE \cite{mae} pre-trained Vision Transformer (ViT) \cite{vit} image-encoder and a set of prompts that are either points, text, or bounding boxes to mask desired objects. This model is also designed for zero-shot transfer. We adopt two variants : \textbf{PromptSAM} \cite{promptsam} and \textbf{GroundedSAM} \cite{ren2024grounded}. PromptSAM uses FocalDINO \cite{focalnet,dino} to generate bounding box proposals as prompts. GroundedSAM uses GroundingDINO \cite{groundingdino} to generate open-vocabulary bounding boxes as prompts to SAM. 
Because GroundingDINO cannot generate discriminate prompts for ``stuff'' categories for semantic segmentation tasks, we only evaluate on instance segmentation.  
\noindent
\textbf{InternImage} \cite{wang2023internimage} is a large-scale CNN-based foundation model which uses deformable convolution as its core operator. It reduces the strict inductive bias of traditional CNNs and makes it possible to learn stronger and more robust patterns with large-scale parameters from massive data like ViTs. 
For our experiments, we use InternImage-XL with cascade method for MS COCO-P and InternImage-H with UperNet framework for ADE20K-P evaluation due to availability of code and weights. 

\noindent
\textbf{Painter} \cite{Painter} is a generalist model which implements in-context learning \cite{gpt3} in NLP to vision tasks. They redefine the output format of chosen tasks to image format and use a masked autoencoder based approach for training the model. Painter has achieved competitive performance compared to well-established task-specific models, which makes it a very powerful model for our task.

\begin{table*}[t!]
    \centering
    \caption{\textbf{Architectural details of the models} used for robustness analysis and their size and the number of parameters used for fine-tuning (FT) the model on the evaluation datasets. Here, Param-M: model parameters in millions, T-Param-M: trainable model parameters in millions, FT-C: finetuned on COCO, FT-A: finetuned on ADE20K}
    \label{tab:model_summary}
    \resizebox{.99\linewidth}{!}{\begin{tabular}{p{.2cm}|l|l|r|r|c|c}
    \hline
    & Model & Backbone(s) & Param-M & T-Param-M & FT-C & FT-A  \\
    \hline
    \multirow{5}{*}{\rotatebox[]{90}{Unimodal}} & 
    MaskDINO \cite{maskdino} & R50\cite{resnet} & 53.25 & 53.25 & True & True \\
    &MaskDINO \cite{maskdino} & SwinL\cite{swin} & 224.39 & 224.39 & True & True \\
    &Mask2Former \cite{mask2former} & R50\cite{resnet} & 44.00 & 44.00 & True & True \\
    &Mask2Former \cite{mask2former} & SwinL\cite{swin} & 216 & 216 & True & True \\
    %Vit-Adpater [] & Vit-Adapter-L\cite{swin} & 348 & 348 & True & True & False\\
    &Vit-Adpater-L \cite{vit-adapter} & ViT\cite{vit} & 348 & 348 & True & False \\
    \hline
    \multirow{5}{*}{\rotatebox[]{90}{Multimodal}}
    &InternImage \cite{wang2023internimage} & InternImage-XL\cite{swin} & 387 & 387 & True & False \\
    %InternImage [] & InternImage-H\cite{swin} & 1120 & 1120 & False & True & False\\
    &Painter \cite{Painter} & ViT\cite{vit} & 371 & 371 & True & True \\
    &PromptSAM \cite{promptsam,segmentanything} & FocalDINO\cite{focalnet,dino},MAE+ViT & 321.86 & 228.12 & True & \textunderscore  \\
    % GT+SAM \cite{segmentanything} & MAE\cite{mae},ViT\cite{vit} & 641.1 & 0 & \textunderscore & \textunderscore & False\\
    &GroundedSAM \cite{groundingdino,segmentanything} & GroundingDINO\cite{groundingdino},MAE\cite{mae}+ViT\cite{vit} & 834.99 & 232.90 & True & False  \\ % GroundingDINO swinB is trained on COCO,O365,GoldG,Cap4M,OpenImage,ODinW-35,RefCOCO, GroundingDINO has 232903808 trainable parameters
    &ODISE \cite{odise} & Mask2Former\cite{mask2former}, CLIP\cite{clip}, GLIDE\cite{glide} & 1,521.90 & 28.10 & True & False  \\
    \hline
    \end{tabular}}
    %\vspace{-5pt}
\end{table*}

\noindent
We compare these 5 multimodal VFM based approaches to 3 unimodal based methods. 

%\vspace{-1.2pt}
\noindent
\textbf{MaskDINO} modifies DINO \cite{dino}, a self-supervised approach using self-distillation. 
ResNet50 (R50) \cite{resnet} and SwinL \cite{swin} backbones are used in our evaluation for MaskDINO. 

\noindent
\textbf{ViT-Adapter} is a pre-training free additional network that can efficiently adapt the plain ViT \cite{vit} to downstream dense prediction tasks without modifying its original architecture. 

\noindent
\textbf{Mask2Former} is a modified MaskFormer \cite{maskformer}, which uses a transformer-based module that produces per-segment embeddings and a pixel-decoder module that produces per-pixel embeddings. 
The pixel-decoder module uses a backbone of either ResNet50 or SwinL.
Table \ref{tab:model_summary} presents more details about all of our used models.
% %\vspace{-.4cm}
\subsection{Datasets}
We use two segmentation benchmark datasets for our experiments: MS COCO Panoptic \cite{coco} and ADE20K \cite{ade20k}. MS COCO dataset has 80 ``things'' categories and 53 ``stuff'' categories. ADE20K has 100 ``things'' and 50 ``stuff'' categories. For each dataset, we perturb an image with each of the 17 corruptions and 5 severities, resulting in 425,000 images for the \textbf{MS COCO-P} dataset and 170,000 images for the \textbf{ADE20K-P}.

\subsection{Benchmark Evaluation Metrics}
\textbf{Performance Metrics:} We evaluate the models on instance and semantic segmentation tasks for our MS COCO-P and ADE20K-P dataset. Each datasets has a category of "things" and "stuff" categories; in which "things" are countable objects like people, animals, etc; whereas "stuff" are amorphous regions like sky, grass, etc. 
\textit{Semantic segmentation} is evaluated on both the ``things'' and ``stuff'' categories using mean intersection over union (mIoU). \textit{Instance segmentation} is only evaluated on the ``things''category using mean average precision (mAP) on the ``things" categories. 
For models trained on panoptic segmentation, all masks assigned to one "thing" category are merged into a single mask. 

\noindent
\textbf{Robustness Metrics:} To measure robustness we use two metrics: absolute and relative robustness \cite{schiappa2023large}. We start by measuring the performance of a trained model $f$ on a clean set of data $A^f_c$ and a corrupted data $A^f_{p,s}$. Here, $A^f_{p,s}$ is corrupted by perturbation $p$ at each severity level $s$. Relative robustness ($\gamma^r_{p,s}$) measures the relative drop in performance between original samples and a corrupted sample, whereas, absolute robustness $\gamma^a_{p,s}$ measures the absolute drop in performance. These can be computed as eqn. \ref{eq:1} and eqn. \ref{eq:2}. 
\begin{align}
    \gamma^r_{p,s} &= 1 - \frac{A^f_c - A^f_{p,s}}{A^f_c} \label{eq:1}\\
    \gamma^a_{p,s} &= 1- \frac{A^f_c - A^f_{p,s}}{100} \label{eq:2}
\end{align}
Because $\gamma^a$ will also depend on the models performance on clean images, we emphasize $\gamma^r$, focusing on the change in performance. Nevertheless, the results of $\gamma^a$ is reported in the supplementary. These metrics are calculated between $0-1$, where higher score means more robustness.

\noindent
\textbf{Implementation Details:} All models are used in accordance to the provided code and model weights. The models Mask2Former, ODISE and MaskDINO all rely on Detectron2 \cite{detectron2} evaluation code. For SAM evaluation, the package \textit{mmsegmentation} \cite{mmseg} was used . 
For GroundedSAM, we were unable to replicate results for its bounding box detector GroundingDINO. While we do provide results for this model, please note that we did our best to replicate given there was no evaluation code provided for either datasets. On the ADE20K dataset, all ODISE and SAM-based models are evaluated zero-shot whereas other models are trained. 

\subsection{Results}
The relative robustness $\gamma^r$ and absolute robustness $\gamma^a$ scores for \textit{instance segmentation} and \textit{semantic segmentation} on MS COCO-P and ADE20K-P is shown respectively in Table \ref{tab:foundation_vs_nonfoundation_mscoco} and Table \ref{tab:foundation_vs_nonfoundation_ade}, where each row corresponds to the average robustness across all corruptions and severity. Here, robustness on each category of segmentation is reported for only those models that had publicly available weights and code for the selected dataset for the selected task. Additional results across all perturbation category for both $\gamma^r$ and $\gamma^a$ is reported in the supplementary.

\begin{table}[t!]
% %\vspace{-15pt}
    \caption{\textbf{Absolute ($\gamma^a$) and Relative ($\gamma^r$) robustness scores for MS COCO-P}, where higher values mean more robust, averaged across all corruptions and severity. Here, \textit{IS} and \textit{SS} denotes instance and semantic segmentation respectively.}
    \label{tab:foundation_vs_nonfoundation_mscoco}
    \resizebox{\linewidth}{!}{\begin{tabular}{l|cc|cc}
        \hline
         & \multicolumn{2}{c|}{IS} & \multicolumn{2}{c}{SS} \\
         \cline{2-5}
         & $\gamma^a$ & $\gamma^r$ & $\gamma^a$ & $\gamma^r$ \\
        \hline
        Mask2Former+R50   &                   $0.86$ &                   $0.68$ &                     $0.85$ &                     $0.75$ \\
        MaskDINO+R50      &                   $0.86$ &                   $0.68$ &                         $0.85$ &                         $0.74$ \\
        Mask2Former+SwinL &                   $0.91$ &                   $0.81$ &                     $0.94$ &         $0.92$ \\
        MaskDINO+SwinL    &                   $0.91$ &       $0.81$ &                         $\mathbf{0.95}$ &                        $\mathbf{0.92}$ \\
        ViT-adapter-L      &                   $0.91$ &                   $0.80$ &                         -- &                         -- \\
        \hline
        ODISE+Label       &                   $0.90$ &                   $0.79$ &            $0.92$ &            $0.88$ \\
        ODISE+Caption & -- & -- & $0.93$ & $0.87$\\
        Prompt+SAM        &                   $0.92$ &                   $0.81$ &                         -- &                         -- \\
        InternImage-XL  &                   $0.91$ &          $0.81$ &                         -- &                         -- \\
        PAINTER           &          $\mathbf{0.95}$ &                   $\mathbf{0.82}$ &                         $0.92$ &                         $0.87$ \\
        GroundedSAM+SwinB &       $0.92$ &                   $0.80$ &         -- &                     -- \\
        \hline
\end{tabular}}
\end{table}

\begin{table}[t!]
% %\vspace{-15pt}
    \caption{\textbf{Absolute ($\gamma^a$) and Relative ($\gamma^r$) robustness scores for ADE20K-P}, where higher values mean more robust, averaged across all corruptions and severity. Here, \textit{IS} and \textit{SS} denotes instance and semantic segmentation respectively.}
    \label{tab:foundation_vs_nonfoundation_ade}
    \resizebox{\linewidth}{!}{\begin{tabular}{l|cc|cc}
        \hline
         & \multicolumn{2}{c|}{IS} & \multicolumn{2}{c}{SS} \\
         \cline{2-5}
         & $\gamma^a$ & $\gamma^r$ & $\gamma^a$ & $\gamma^r$ \\
        \hline
        Mask2Former+R50   &                   $0.89$ &                   $0.57$ &                     $0.84$ &                     $0.65$ \\
        MaskDINO+R50      &                   -- &                   -- &                         $0.82$ &                         $0.63$ \\
        Mask2Former+SwinL &                   $0.94$ &                   $\mathbf{0.92}$ &                     $0.93$ &         $0.87$ \\
        ViT-adapter-L      &                   -- &                  -- &                         $0.94$ &                        $0.89$ \\
        \hline
        ODISE+Label       &                   $\mathbf{0.97}$ &                   $0.79$ &            $\mathbf{0.97}$ &            $\mathbf{0.89}$ \\
        InternImage-H   &       -- &         -- &                         $0.93$ &                         $0.87$ \\
        PAINTER           &         -- &                   -- &                         $0.92$ &                         $0.83$ \\
        GroundedSAM+SwinB &       $0.95$ &                   $0.73$ &         -- &                     -- \\
        \hline
\end{tabular}}
%\vspace{-8pt}
\end{table}

\section{Analysis and Discussion}
\label{sec:analysis}
\begin{figure*}[t!]
    \centering
    \includegraphics[width=\linewidth]{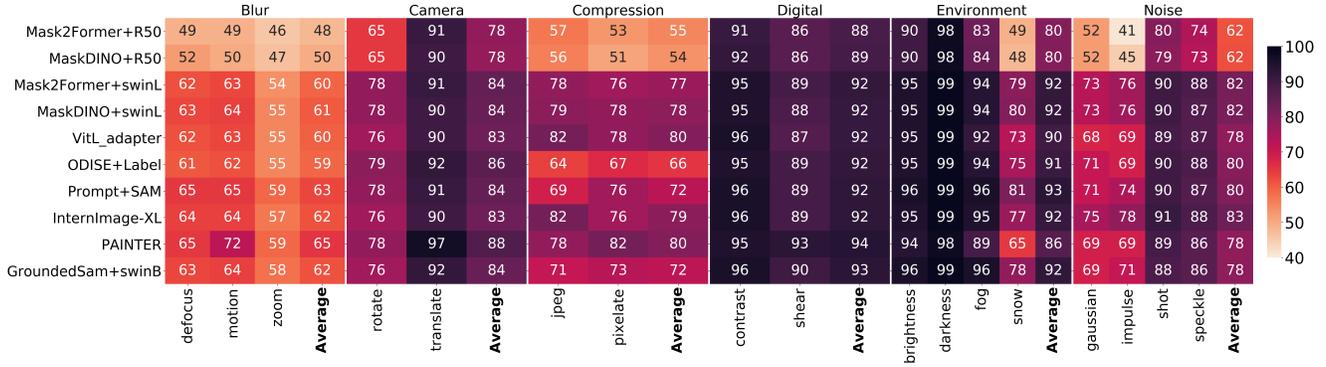}
    \caption{\textbf{Relative robustness score $\gamma^r$ on instance segmentation} for the MS COCO-P dataset. Here the Y-axis denotes the models we evaluated and the x-axis denotes $\gamma^r$ for each corruption averaged over severity.}
    \label{fig:coco_distribution_shifts_heatmap}
    %\vspace{-5pt}
\end{figure*}

\begin{figure}[t!]
    \centering
    \includegraphics[width=\linewidth, height=5cm]{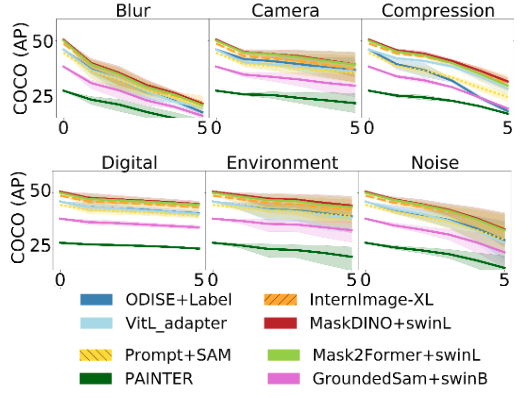}
    \caption{\textbf{mAP score on instance segmentation on MS COCO-P}. Here x axis denotes severity ranging from 0 (no corruption) to 5 (severe corruption) and y axis denotes the model performance measured by mean average precision (mAP).}
    \label{fig:coco_distribution_shifts}
    %\vspace{-10pt}
\end{figure}

\textbf{All models struggle with blur and compression:}
The relative robustness $\gamma^r$ scores and mean average precision (mAP) scores  for \textit{instance segmentation} for MS COCO-P for all distribution shifts is shown respectively in Figure \ref{fig:coco_distribution_shifts_heatmap} and Figure \ref{fig:coco_distribution_shifts}. 
\begin{figure*}[t!]
    \centering
    \includegraphics[width=\linewidth]{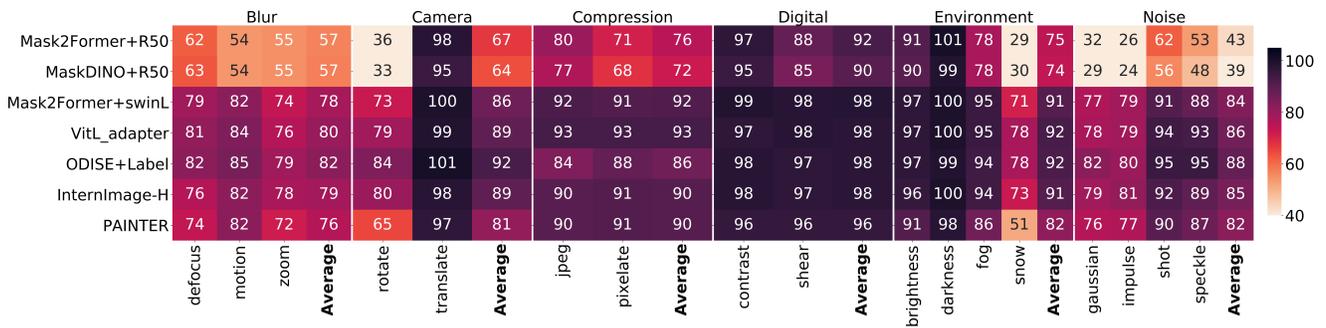}
    \caption{\textbf{Relative robustness score $\gamma^r$ on semantic segmentation on ADE20K-P}. Here, the Y-axis denotes the models we evaluated and x-axis denotes $\gamma^r$ for each corruption averaged over severity.}
    \label{fig:ade20k_semantic_heatmap}
\end{figure*}
We observe that the selected models are typically robust to all shifts with the exception to blur and compression. Figure \ref{teaser} shows an example of compression corruptions for ODISE and GroundedSAM. Here, even for compression at severity level 5, the objects are clearly visible for the human eye. However, ODISE is struggling to properly classify objects even at severity 1 (person is denoted as handbag too), even though the generated mask is correct. However, as the severity increases, even though the object boundary is clear to the naked human eye, ODISE's generated mask gets more degraded in quality. GroundedSAM on the other hand generates correct masks and accurate recognition even at severity 5. 

\noindent
In summary, while all models struggle with \textit{blur and compression} corruptions, ODISE and SAM are the particularly lower performing ones in terms of robustness. Since both of these foundation models use a generative model for image embeddings, this could be a reflection of generative model robustness to compression based corruptions. Figure \ref{fig:coco_distribution_shifts} presents the performance of all models in terms of mean Average Precision (mAP). Interestingly, the model with the highest robustness score, PAINTER, exhibits the lowest mAP. Despite its lower mAP compared to other models, PAINTER demonstrates superior robustness in terms of $\gamma^r$ score across various corruption categories. This highlights a trade-off between selecting a model based on its performance versus its robustness. In this case, PAINTER's resilience to corruption shifts makes it a compelling choice despite its slightly lower overall performance in mAP.

\begin{figure}[t!]
    %\vspace{-10pt}
    \centering
    \includegraphics[width=\linewidth]{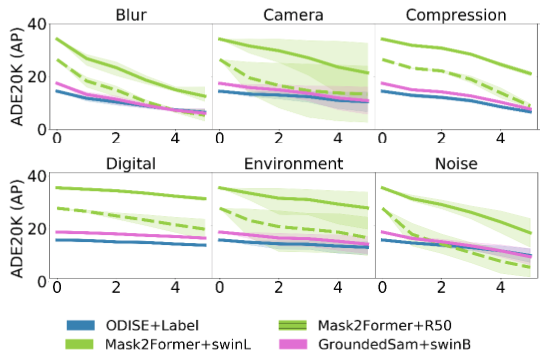}
    \caption{\textbf{mAP score on instance segmentation on ADE20K-P}. Here x axis denotes severity and y axis denotes the model performance measured by mean average precision (mAP). All ODISE and SAM-based models are evaluated zero-shot.}
    \label{fig:ade20k_distribution_shifts_instance}
    %\vspace{-8pt}
\end{figure}

\noindent
\textbf{Multimodal models are not typically more robust or higher performing; but are consistent on zero-shot:}
Since traditionally ADE20K is more popularly used for semantic segmentation than instance segmentation, we report the relative robustness $\gamma^r$ scores for ADE20K-P dataset across all categories for semantic segmentation in Figure \ref{fig:ade20k_semantic_heatmap}. 
In Table \ref{tab:foundation_vs_nonfoundation_mscoco}, Table \ref{tab:foundation_vs_nonfoundation_ade}, Figure \ref{fig:coco_distribution_shifts_heatmap} and Figure \ref{fig:ade20k_semantic_heatmap}, we show 
robustness results across model types defined by whether a model is a "foundation" model or "non-foundation" model, and whether the non-foundation model uses a CNN or a transformer based backbone. Now, 
these results do not provide convincing evidence that all multimodal models are typically more robust than unimodal. However, the most robust model is indeed a multimodal model, but this does not necessarily prove multimodal models' strength over unimodal models. 
There is a more noticeable difference for absolute robustness (Table \ref{tab:foundation_vs_nonfoundation_ade}) on the ADE20K-P datset where foundation models are evaluated zero-shot. So while models may not be typically more robust or higher performing, their zero-shot capability allows for much greater flexibility. When observing Figure \ref{fig:ade20k_distribution_shifts_instance}, we see that even for hard cases of corruption like blur and compression, even as the severity increases, the zero shot performance of the multimodal models remain relatively stable in comparison to unimodals. This same observation is seen in case of robustness as well (Figure \ref{fig:ade20k_semantic_heatmap}).This consistent performance of multimodal models is seen across all different categories of corruption. In summary, while the selected multimodal models are not typically more robust or higher-performing than the unimodal ones, they show promising zero-shot capabilities that have competitive robustness scores across both instance and semantic segmentation tasks.

\noindent
\textbf{All models lack robustness in texture non-preserving corruptions:}
Even though as per Figure \ref{fig:ade20k_semantic_heatmap}, we observe that models show robustness across the corruption category as well for the ADE20K-P dataset; from Figure \ref{fig:coco_distribution_shifts_heatmap}, we see almost all models performance are affected by all corruptions in the blur, compression category; snow from the environment category, gaussian, and impulse from the noise category of corruptions for the MS COCO-P dataset. Apart from compression, robustness drop for corruptions due to all the aforementioned categories are also valid in case of the ADE20K-P dataset. These results align with previous works \cite{kamann2020benchmarking, geirhos2018imagenet} that distortions that corrupt the texture of an image have a negative effect on model robustness compared to texture-preserving corruptions such as brightness, contrast, and geometric corruption. This shows that both multimodal and unimodal models are not robust to distortions that corrupts image texture. We have finetuned a few models on an augmented dataset consisting of these texture non-preserving corruption to see how it affects performance. 

\begin{figure}
    \centering
    \includegraphics[width=\linewidth,height=3cm]{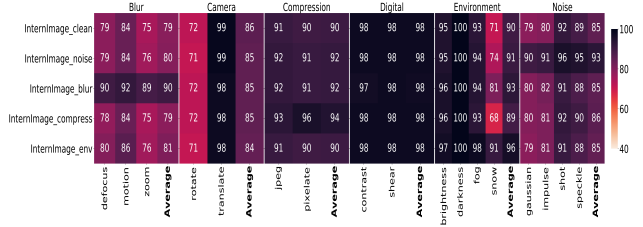}
    \caption{\textbf{Fine-tuned performance of InternImage on semantic segmentation for the ADE20K-P dataset}. Y-axis: Augmentation used for fine-tuning (expect first row). X-axis: model Relative Robustness score for each corruption averaged over severity.}
    \label{fig:Internimage_finetune}
    %\vspace{-.3cm}
\end{figure}

\noindent
In this experiment, we fine-tune the models InternImage-XL, Vit-adapter-L, and Mask2Former+swinL using an augmented dataset specific to a particular category. The objective is to assess whether these augmentations enhance the models' robustness. We evaluate the models on the ADE20K-P dataset for semantic segmentation task, focusing on their performance under varied perturbations. We present the results of the InternImage model in Figure \ref{fig:Internimage_finetune}, whereas results of other models on the augmented dataset and more details about the fine-tuning dataset is provided in the supplementary. Across all models, augmentations generally elevate the perturbation score of their respective category, except for compression augmentation in Mask2Former, which adversely affects performance across all categories, including compression. Notably, blur and noise augmentation substantially elevate robustness scores in each model's relevant category. 

\begin{figure*}[t!]
    %\vspace{-5pt}
    \centering
    \includegraphics[width=\linewidth]{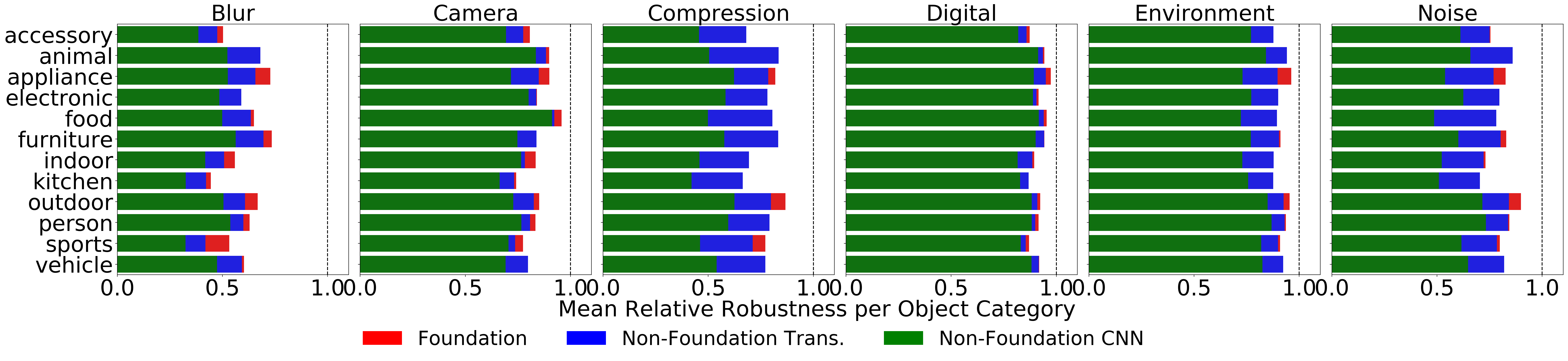}
    \caption{\textbf{Relative robustness scores $\gamma^r$ across object super categories and corruptions} categories on MS COCO-P for multimodal models and unimodal models. The higher $\gamma^r$ the more relatively robust.}
    \label{fig:coco_per_object}
\end{figure*}

\begin{table*}[t!]
\setlength{\tabcolsep}{3pt}
\caption{\textbf{Mean Relative Robustness ($\gamma^r$) scores for object super categories} for MS COCO-P dataset. A higher $\gamma^r$ score is more robust with the top score in bold and second underlined. }
%\vspace{-2mm}
\label{tab:per_object_rel_robustness_foundation}
    \resizebox{\linewidth}{!}{\begin{tabular}{l|cccccccccccc}
\toprule
 &           accessory &              animal &           appliance &          electronic &                food &           furniture &              indoor &             kitchen &             outdoor &              person &              sports &             vehicle \\
\hline
Unimodal CNN    &              $0.62$ &              $0.71$ &              $0.65$ &              $0.68$ &              $0.65$ &              $0.68$ &              $0.61$ &              $0.58$ &              $0.72$ &              $0.74$ &              $0.63$ &              $0.68$ \\
Unimodal Trans. &  $\underline{0.74}$ &     $\mathbf{0.86}$ &  $\underline{0.81}$ &     $\mathbf{0.80}$ &     $\mathbf{0.82}$ &  $\underline{0.83}$ &  $\underline{0.74}$ &     $\mathbf{0.71}$ &  $\underline{0.82}$ &  $\underline{0.82}$ &  $\underline{0.74}$ &     $\mathbf{0.81}$ \\
Multimodal            &     $\mathbf{0.74}$ &  $\underline{0.83}$ &     $\mathbf{0.87}$ &  $\underline{0.77}$ &  $\underline{0.81}$ &     $\mathbf{0.84}$ &     $\mathbf{0.76}$ &  $\underline{0.71}$ &     $\mathbf{0.86}$ &     $\mathbf{0.82}$ &     $\mathbf{0.78}$ &  $\underline{0.80}$ \\

\hline
\end{tabular}}
\end{table*}

\noindent
\textbf{Multimodal models are relatively more robust to certain objects; especially under blur and compression:}
To evaluate how model performance per object is impacted under different corruptions, we evaluate per-object relative robustness ($\gamma^r$) scores. Figure \ref{fig:coco_per_object} shows a summary of the $\gamma^r$ across 11 super-categories for objects and corruptions under each distribution shift category for the MS COCO-P dataset. For mapping of object to super category, original MS COCO documentation is followed. 

\noindent
When looking at super-categories in Figure \ref{fig:coco_per_object}, we observe that multimodal models demonstrate greater relative robustness for certain object categories for certain perturbations. This is most noticeable under compression, blur and noise for objects in ``outdoor'' and ``sports''. To better understand these patterns, Table \ref{tab:per_object_rel_robustness_foundation} reports the average relative robustness ($\gamma^r$) scores of each object super category across all corruptions and severity. From this table, we observe the selected multimodal models are typically more robust against objects under ``appliance'', ``furniture'', ``outdoor'' and ``sports'' when averaged across all corruptions and severity. While objects in furniture and outdoor tend to be quite large, the objects in sports are quite small. Therefore the size of the objects may not be the factor for this robustness. However, why multimodal models show more robustness across these specific object super categories need more exploration. One area of exploration could be the open-vocabulary training paradigm of the multimodal models where these models have been exposed to a broader and more diverse set of labels and descriptions of the aforementioned categories, enabling them to generalize better across various contexts and conditions. This wide exposure helps the models to learn more robust and transferable features that can be effective across different categories, even under distortions or corruptions. Nevertheless, this area needs a lot more exploration to understand the proper reasons behind this higher performance in certain object categories.

\begin{figure}
    \centering
    \includegraphics[width=\linewidth]{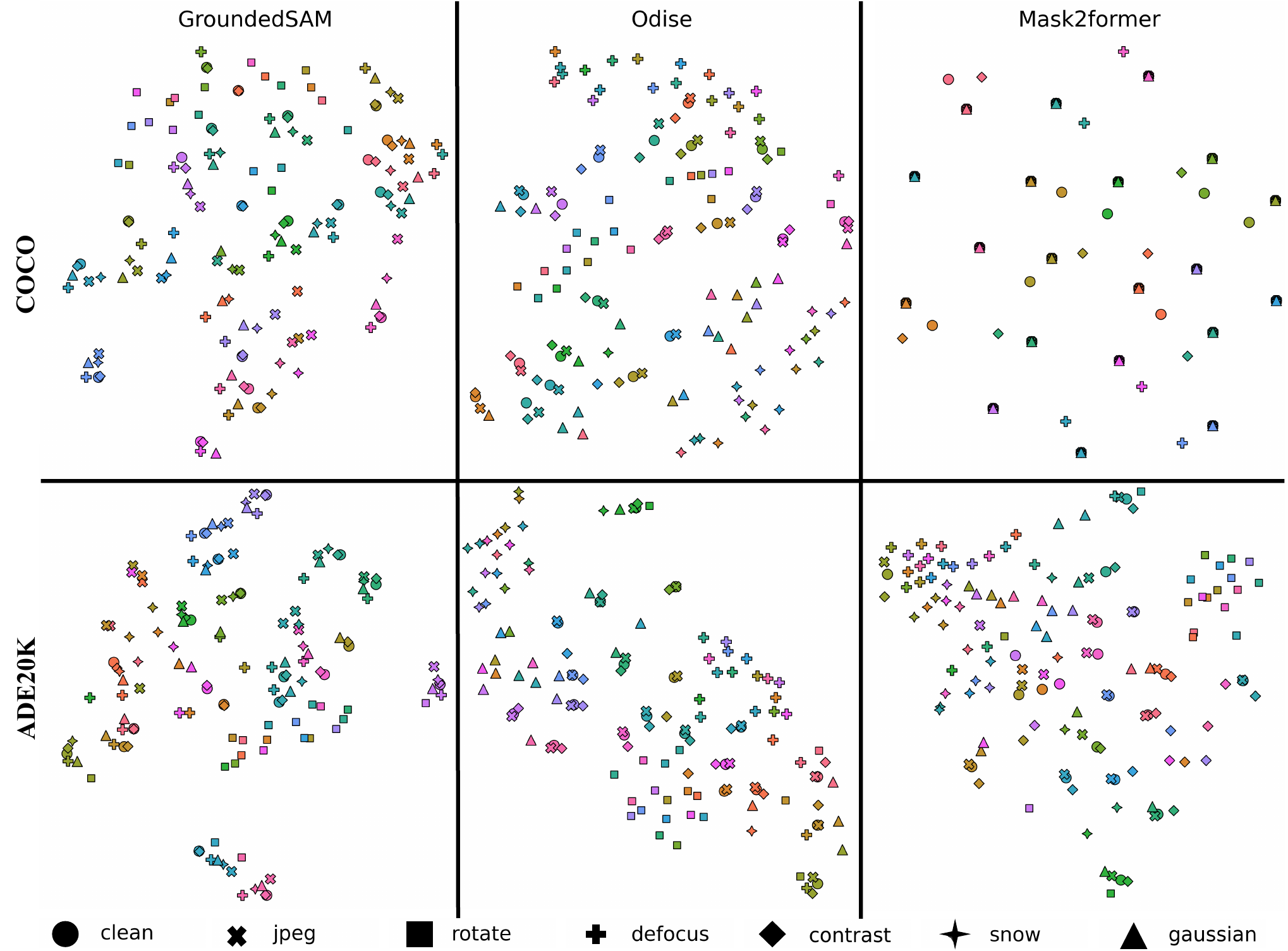}
    \caption{\textbf{Visualization of feature spaces of image encoders} from multimodal ODISE and GroundedSAM and unimodal Mask2Former. For a subset of images in MS COCO-P, we extract multiple variations under different corruptions at severity 3. Each color is a single image, while marker shape is corruption type.}
    \label{fig:tsne}
    %\vspace{-.7cm}
\end{figure}

\noindent
\textbf{The more similar corrupted image features are to original, likely more robust:}
Figure \ref{fig:tsne} shows TSNE visualizations of feature spaces for image encoders from multimodal model ODISE, GroundedSAM and unimodal Mask2Former. This helps us observe whether models encode an image in the same space as its corrupted versions, clustering by image, or if it clusters by corruption type. When observing at MS COCO-P, we see that the unimodal model, Mask2Former, clusters representations by image as indicated by the overlap of different corruptions that are close to the original image. GroundedSAM seems to also be clustering by image. ODISE, on the other hand, does some clustering based on the image, but with more noticeable clustering by corruption type. 

\noindent
When evaluated on ADE20K-P datasets, the clustering tendency of GroundedSAM and ODISE remained the same. However, for Mask2Former, the tight clustering based on images that was observed in case of MS COCO-P is no longer present and in this case the clustering is more often based on corruption type. This aligns with robustness as well, Mask2former is more robust typically on MS COCO-P while noticeably less robust on ADE20K-P (Table \ref{tab:foundation_vs_nonfoundation_mscoco}, Table \ref{tab:foundation_vs_nonfoundation_ade}). This may indicate that an additional measurement of robustness is the more similar corrupted versions of an image are to the original in latent space, the more robust.
%\vspace{-.3cm}

\section{Conclusion}
\label{sec:conclusion}
In this benchmark, we evaluated multimodal Visual Foundation Models (VFMs) and unimodal models for segmentation on MS COCO-P and ADE20K-P datasets which are perturbed using 17 categories of corruption that reflect real-world data corruptions across 5 different level of severity. Our study provides several interesting insights about the selected models. (1) All selected models struggle with blur and and compression based corruptions (2) Although multimodal VFMs are not noticeably more robust than unimodal models however, they show competitive robustness results when evaluated zero-shot. (3) selected multimodal VFMs show higher relative robustness for specific object-types like those found in sports, outdoor and appliance compared to other unimodal models.
We hope these findings and the benchmark in this work can potentially open up interesting questions about robustness segmentation and foundation models. 

{
    \small
    \bibliographystyle{ieeenat_fullname}
    \bibliography{main}
}

\clearpage
\setcounter{page}{1}
\setcounter{section}{0}
\setcounter{table}{0}
\setcounter{figure}{0}
\maketitlesupplementary

The supplementary will provide more details about the benchmark datasets, models, and additional results. In Section \ref{supp:perturbations}, we provide more details on how our real-world perturbations were generated for the MS COCO-P and ADE20K-P dataset. In Section \ref{supp:additional_results} we provide additional results from our benchmark. 

\section{Distribution Shift Perturbations}
\label{supp:perturbations}
\begin{figure*}
    \centering
    \includegraphics[width=\linewidth]{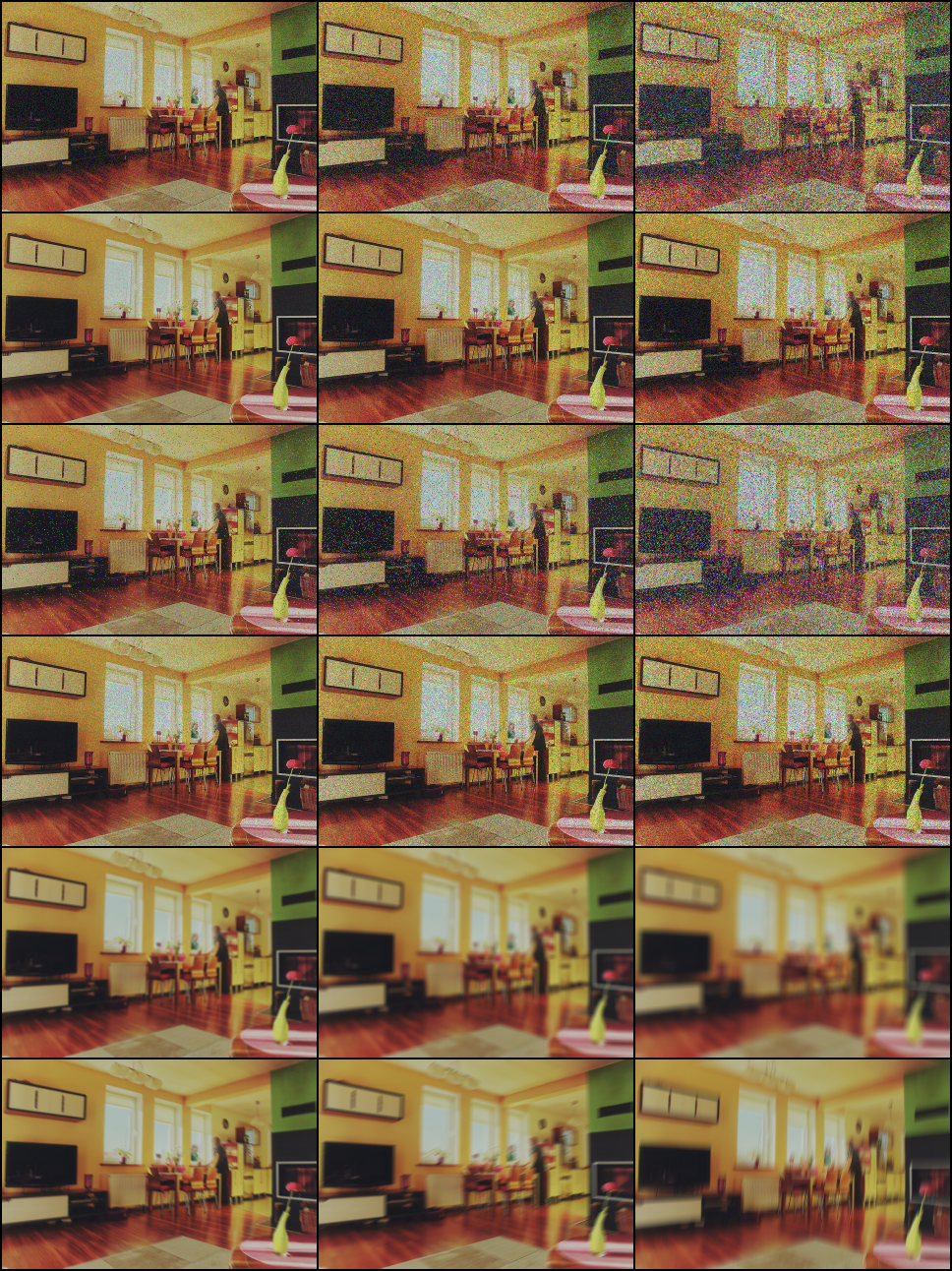}
    \caption{\textbf{Visual example from the \textbf{MS COCO-P} dataset for perturbations \textit{gaussian, shot, impulse, speckle, defocus, motion}} across 1, 3, 5 severity.}
    \label{fig:all_perturbation1}
\end{figure*}

\begin{figure*}
    \centering
    \includegraphics[width=\linewidth]{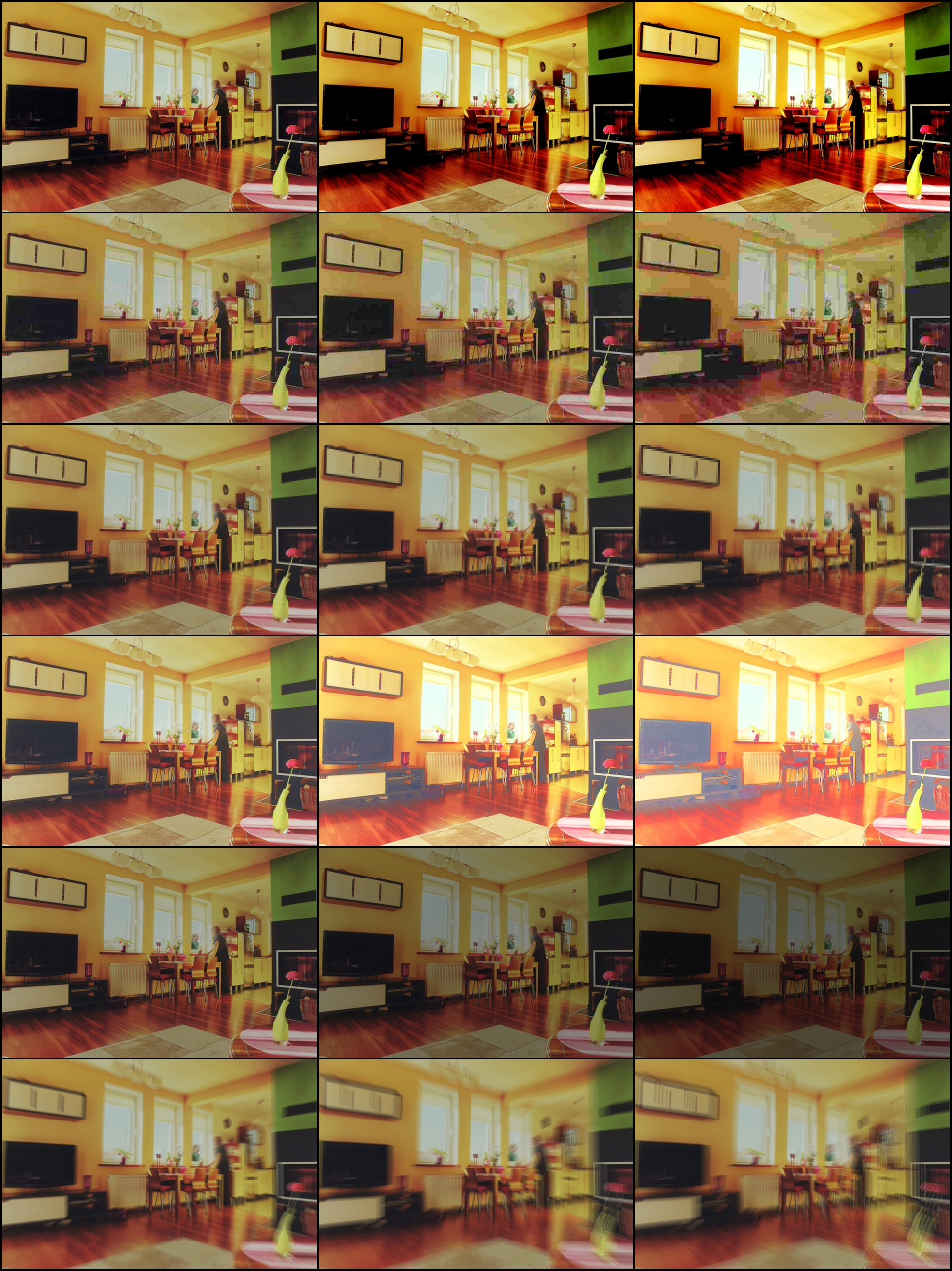}
    \caption{\textbf{Visual example from the \textbf{MS COCO-P} dataset for perturbations \textit{contrast, jpeg, pixelate, brightness, darkness, zoom}} across 1, 3, 5 severity.}
    \label{fig:all_perturbation2}
\end{figure*}

\begin{figure*}
    \centering
    \includegraphics[width=\linewidth]{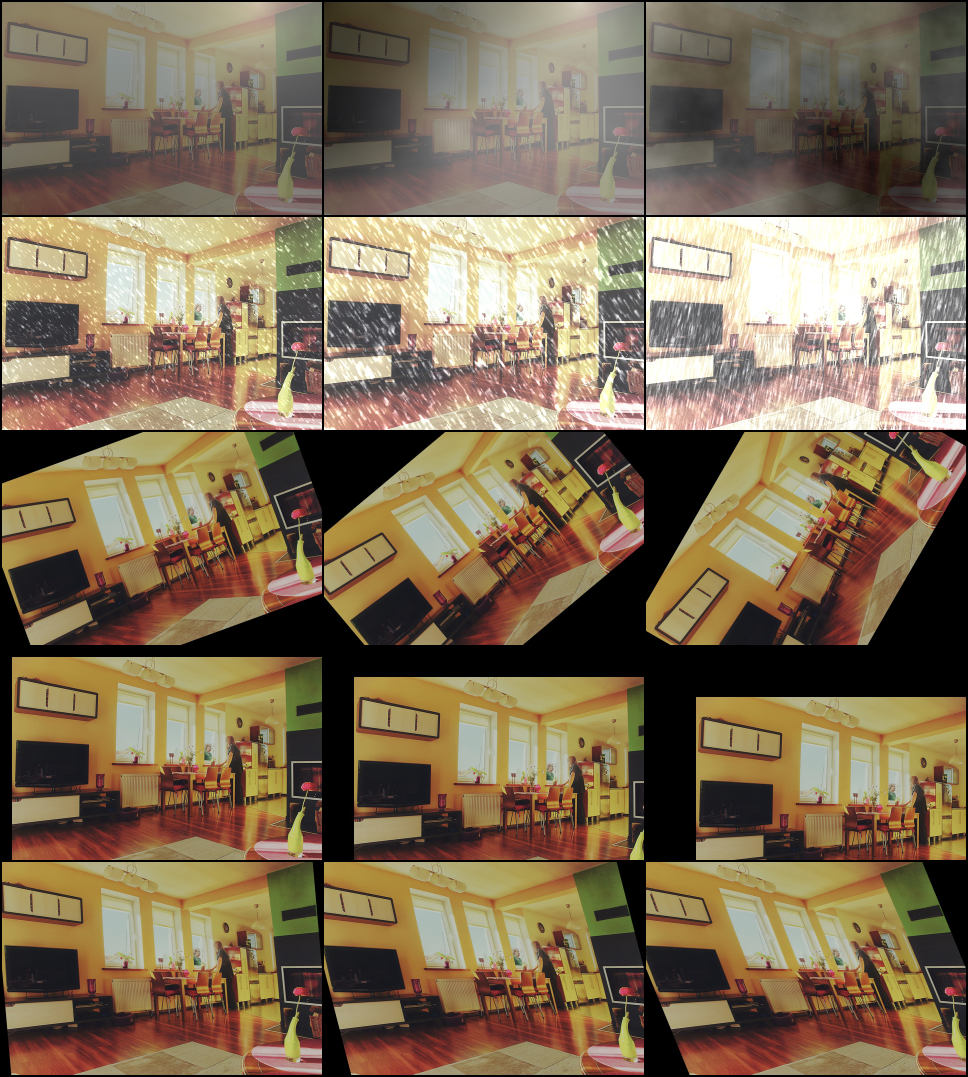}
    \caption{\textbf{Visual example from the \textbf{MS COCO-P} dataset for perturbations \textit{fog, snow, rotate, translate, shear}} across 1, 3, 5 severity.}
    \label{fig:all_perturbation3}
\end{figure*}

We used 17 types of algorithmically generated corruptions to generate a perturbed dataset. These corruption are from different categories like noise, blur, environment, digital and camera. We have given a overview plot for all perturbation in Figure \ref{fig:all_perturbation1},\ref{fig:all_perturbation2},\ref{fig:all_perturbation3}.

\noindent
\textbf{Noise}. We have Gaussian, shot, impulse, and speckle noise in the noise category. Gaussian noise is modeled by adding random values sampled from a Gaussian distribution to the pixel intensities of a clean image, the standard deviation of the Gaussian noise determines the severity. Shot noise is modeled by applying Poisson distribution to the pixel values of the clean image. Impulse noise is modeled by adding salt and pepper noise to the clean image, and the density of the noise determines the severity. Speckle noise is generated by adding a normal noise distribution whose intensities are proportional to the clean image pixel intensities, and the standard deviation of the noise determines the severity.

\noindent
\textbf{Blur}. We have defocus, motion, and zoom blur in the Blur category. Defocus blur is modeled by convolving the clean image with a blur kernel, here the blur kernel is a circular Gaussian blur kernel, and the blur radius determines the severity of noise. Zoom blur is modeled by averaging multiple zoomed images generated by scaling up the image and cropping out of the boundary region to maintain the original shape. Here, the list of scaling factors used determines the severity.

\noindent
\textbf{Compression}. In the digital category, we have jpeg and pixelate corruption. Jpeg corruption is generated by saving the image in jpeg format by reducing the quality, and the quality determines the severity. Pixelate corruption is modeled by upsampling a low-resolution image, and the severity is controlled by how much it was downsampled before upsampling.

\noindent
\textbf{Digital}
Contrast corruption is generated by blending a clean image with another image in which all pixel values are set to the mean value of the clean image. Here the blending factor determines the severity. Shear corruption is generated with the help of imgaug \cite{imgaug}.

\noindent
\textbf{Camera}
In the geometric category, we have translate and rotate. Both translate and rotate are implemented with the help of imgaug \cite{imgaug} library to generate corrupted images and their corresponding annotations.

\noindent
\textbf{Environment}
Darkness corruption is modeled by blending a black image with a clean image with a blending factor determined by severity. We additionally have snow and fog corruption which are algorithmically generated images that try to mimic real-life fog and snow. 

\noindent
Fog, Snow, motion blur, brightness, and shear perturbations are all implemented using imgaug \cite{imgaug} library.

\section{Additional Results}
\label{supp:additional_results}
Here we provide additional results and more details on the robustness scores and performance of the selected models. Absolute Robustness ($\gamma^a$) scores are additionally included here and are the absolute drop in performance while Relative Robustness ($\gamma^r$) is the relative drop based on the original model score. 

\subsection{Instance Segmentation}
\begin{figure*}[t!]
    \centering
    \includegraphics[width=\linewidth]{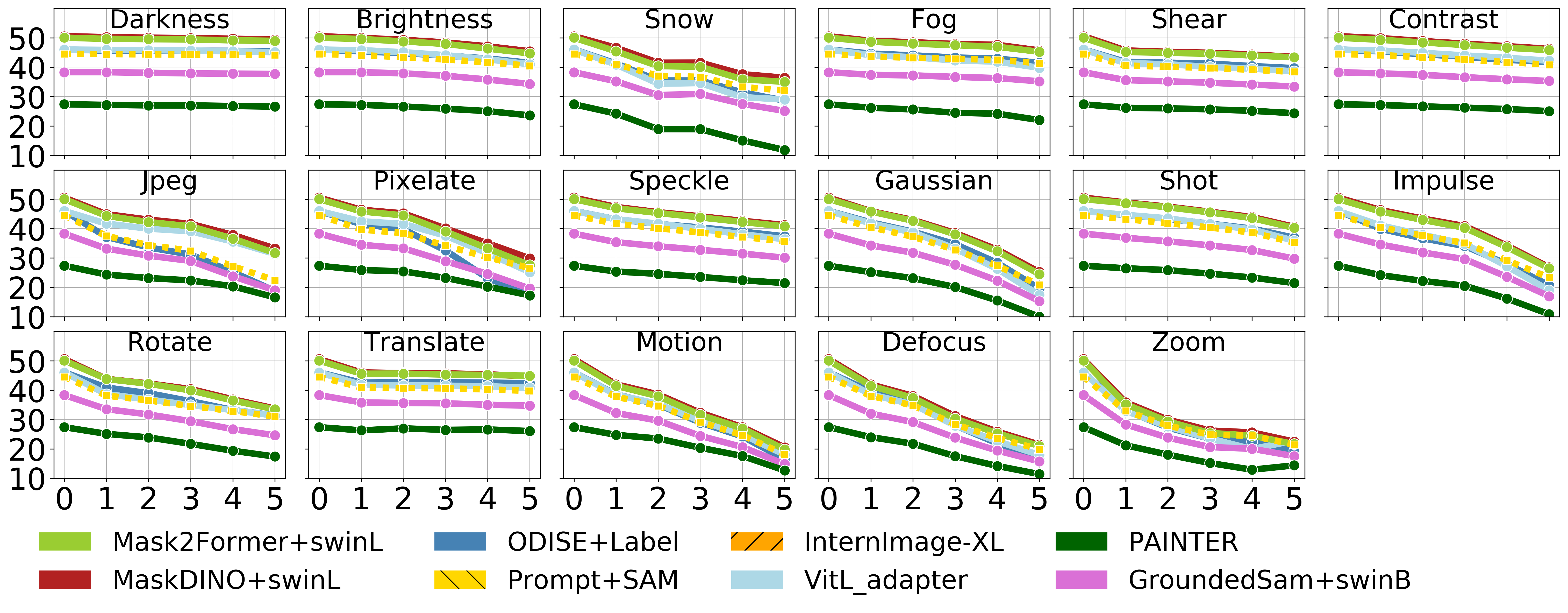}
    \caption{\textbf{Results for each corruption and each severity for instance segmentation measured by average precision (AP) on the MS COCO-P dataset}. x-axis: Severity ranges from 0 (no corruption) to 5 (most corruption). y-axis: AP results for instance segmentation.}
    \label{fig:coco_corruption_line_plots}
    %\vspace{-10pt}
\end{figure*}

\begin{figure*}[t!]
    \centering
    \includegraphics[width=\linewidth]{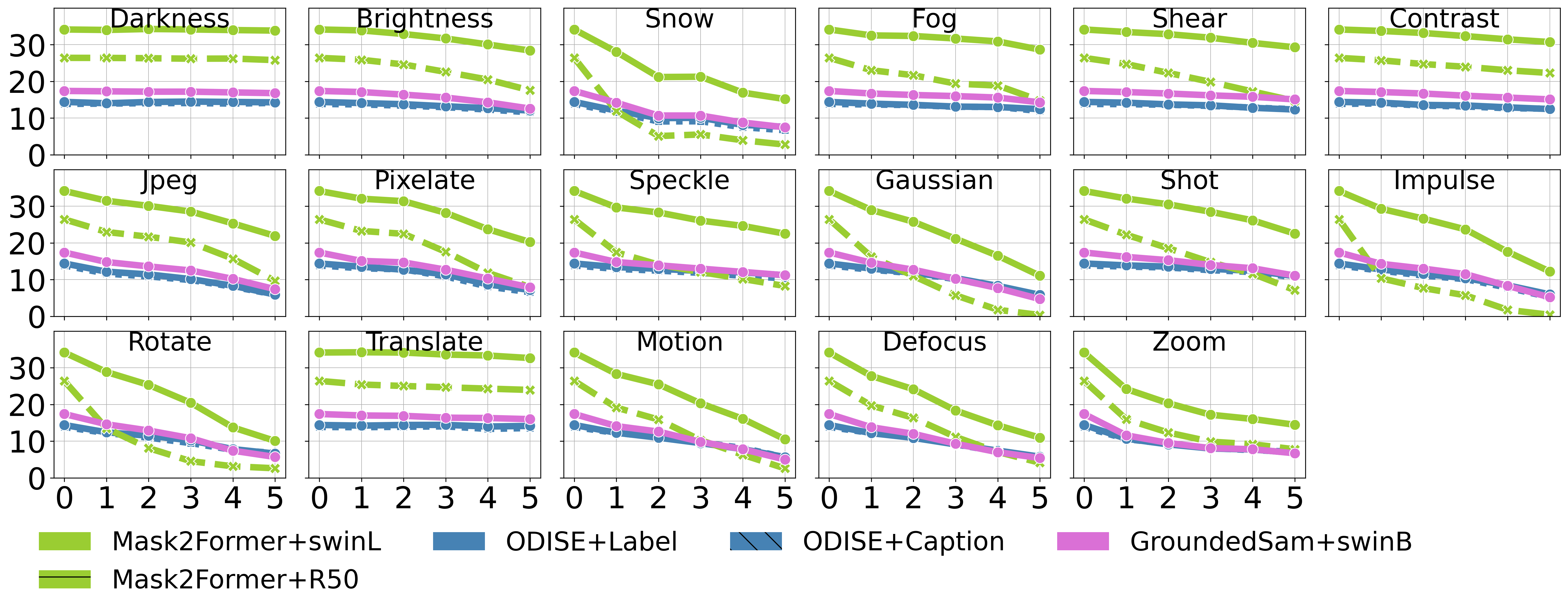}
    \caption{\textbf{Results for each corruption and each severity for instance segmentation measured by average precision (AP) on the ADE20K-P dataset.} x-axis: Severity ranges from 0 (no corruption) to 5 (most corruption). y-axis: AP results for instance segmentation.}
    \label{fig:ade20k_corruption_line_plots}
    %\vspace{-10pt}
\end{figure*}

Table \ref{tab:absolute_instance_results} and Table \ref{tab:relative_results} respectively shows results for absolute robustness scores $\gamma^a$ and relative robustness scores $\gamma^r$ for the selected models. $\gamma^a$ measures the absolute drop in performance as compared to $\gamma^r$ which measures relative drop to original performance of a given model. These results are averaged across severity for each corruption type. One observation is that when comparing results for ADE20K-P where ODISE and SAM are evaluated zero-shot, absolute robustness is much higher than relative. This indicates that while models may start with lower performance overall, they show more consistent results across perturbations. More details on model behavior across severity for instance segmentation are shown in Figure \ref{fig:coco_corruption_line_plots} on MS COCO-P and Figure \ref{fig:ade20k_corruption_line_plots} for ADE20K-P where multimodal models are evaluated on zero-shot. On MS COCO-P, we see very similiar trends across all corruptions except for compression-based. For both JPEG and Pixelate, we see a  some different trends for ODISE showing a sudden drop at severity 3. For ADE20K, where multimodal are evaluated zero-shot, we see more consistent results across severity and more declines from the Mask2Former model. This supports the conclusion that of the selected multimodal models, while their zero-shot performance is low, their absolute robustness across severity is good and performance consistent. Table \ref{tab:per_supercategory_coco_Relative_Robustness} presents the object super-category wise robustness scores for both $\gamma^a$ and $\gamma^r$. We observe that multimodal models are noticeabley more relatively robust in certain object categories.

\begin{table*}
\centering
    \caption{\textbf{Absolute Robustness scores ($\gamma^a$) for instance segmentation} on models on the MS COCO-P and ADE20K-P dataset. Models with the least relative drop in performance are in bold, and models that are second least are underlined.}
    \label{tab:absolute_instance_results}
    \resizebox{\linewidth}{!}{%
    \begin{tabular}{l|cccc|cc|cc|cccc|cc|ccc}
\hline
 \multirow{2}{*}{\textbf{COCO} ($\gamma^a$)} & \multicolumn{4}{c|}{\textbf{Environment}} & \multicolumn{2}{c|}{\textbf{Digital}} & \multicolumn{2}{c|}{\textbf{Compression}} & \multicolumn{4}{c|}{\textbf{Pixel Noise}} & \multicolumn{2}{c|}{\textbf{Camera}} & \multicolumn{3}{c}{\textbf{Blur}} \\
 \cline{2-18}
 &            dark &          bright &                snow &                 fog &               shear &            contrast &                jpeg &            pixel. &             speckle &            gauss. &                shot &             impulse &              rotate &           translate &              motion &             defocus &                zoom \\
 \hline
Mask2Former+R50   &              $0.99$ &              $0.96$ &              $0.78$ &              $0.93$ &              $0.94$ &              $0.96$ &              $0.81$ &              $0.79$ &              $0.89$ &              $0.79$ &              $0.91$ &              $0.74$ &              $0.85$ &              $0.96$ &              $0.78$ &              $0.78$ &              $0.77$ \\
MaskDINO+R50      &              $0.99$ &              $0.96$ &              $0.77$ &              $0.93$ &              $0.94$ &              $0.96$ &              $0.81$ &              $0.78$ &              $0.88$ &              $0.78$ &              $0.91$ &              $0.76$ &              $0.85$ &              $0.96$ &              $0.78$ &              $0.79$ &              $0.76$ \\
Mask2Former+swinL &              $0.99$ &              $0.97$ &              $0.89$ &              $0.97$ &              $0.94$ &              $0.98$ &              $0.89$ &              $0.88$ &              $0.94$ &              $0.87$ &              $0.95$ &              $0.88$ &              $0.89$ &              $0.95$ &              $0.81$ &              $0.81$ &              $0.77$ \\
MaskDINO+swinL    &              $0.99$ &              $0.97$ &              $0.90$ &              $0.97$ &              $0.94$ &              $0.97$ &              $0.90$ &              $0.89$ &              $0.94$ &              $0.87$ &              $0.95$ &              $0.88$ &              $0.89$ &              $0.95$ &              $0.82$ &              $0.81$ &              $0.77$ \\
VitL-adapter      &              $1.00$ &              $0.98$ &              $0.88$ &              $0.96$ &              $0.94$ &              $0.98$ &  $\underline{0.92}$ &              $0.90$ &              $0.94$ &              $0.85$ &              $0.95$ &              $0.86$ &              $0.89$ &              $0.95$ &              $0.83$ &              $0.82$ &              $0.79$ \\
\hline
ODISE+Caption     &     $\mathbf{1.00}$ &              $0.98$ &              $0.89$ &              $0.97$ &  $\underline{0.96}$ &              $0.98$ &              $0.86$ &              $0.87$ &  $\underline{0.95}$ &  $\underline{0.88}$ &  $\underline{0.96}$ &              $0.87$ &  $\underline{0.92}$ &  $\underline{0.98}$ &              $0.86$ &              $0.85$ &              $0.83$ \\
ODISE+Label       &              $1.00$ &              $0.98$ &              $0.88$ &              $0.97$ &              $0.95$ &              $0.97$ &              $0.83$ &              $0.85$ &              $0.94$ &              $0.87$ &              $0.95$ &              $0.86$ &              $0.90$ &              $0.97$ &              $0.83$ &              $0.82$ &              $0.79$ \\
Prompt+SAM        &  $\underline{1.00}$ &              $0.98$ &  $\underline{0.92}$ &  $\underline{0.98}$ &              $0.95$ &              $0.98$ &              $0.86$ &              $0.89$ &              $0.94$ &              $0.87$ &              $0.95$ &              $0.89$ &              $0.90$ &              $0.96$ &              $0.84$ &              $0.84$ &              $0.82$ \\
InternImage-XL    &              $0.99$ &              $0.98$ &              $0.89$ &              $0.98$ &              $0.94$ &              $0.98$ &              $0.91$ &              $0.88$ &              $0.94$ &              $0.88$ &              $0.95$ &  $\underline{0.89}$ &              $0.88$ &              $0.95$ &              $0.83$ &              $0.82$ &              $0.79$ \\
PAINTER           &              $1.00$ &  $\underline{0.98}$ &              $0.90$ &              $0.97$ &     $\mathbf{0.98}$ &     $\mathbf{0.99}$ &     $\mathbf{0.94}$ &     $\mathbf{0.95}$ &     $\mathbf{0.96}$ &     $\mathbf{0.91}$ &     $\mathbf{0.97}$ &     $\mathbf{0.91}$ &     $\mathbf{0.94}$ &     $\mathbf{0.99}$ &     $\mathbf{0.92}$ &     $\mathbf{0.90}$ &     $\mathbf{0.89}$ \\
GroundedSam+swinB &              $1.00$ &     $\mathbf{0.98}$ &     $\mathbf{0.92}$ &     $\mathbf{0.98}$ &              $0.96$ &  $\underline{0.98}$ &              $0.89$ &  $\underline{0.90}$ &              $0.94$ &              $0.88$ &              $0.96$ &              $0.89$ &              $0.91$ &              $0.97$ &  $\underline{0.86}$ &  $\underline{0.86}$ &  $\underline{0.84}$ \\
\hline
\hline
 \multirow{2}{*}{\textbf{ADE20K} ($\gamma^a$)} & \multicolumn{4}{c|}{\textbf{Environment}} & \multicolumn{2}{c|}{\textbf{Digital}} & \multicolumn{2}{c|}{\textbf{Compression}} & \multicolumn{4}{c|}{\textbf{Pixel Noise}} & \multicolumn{2}{c|}{\textbf{Camera}} & \multicolumn{3}{c}{\textbf{Blur}} \\
 \cline{2-18}
 &            dark &          bright &                snow &                 fog &               shear &            contrast &                jpeg &            pixel. &             speckle &            gauss. &                shot &             impulse &              rotate &           translate &              motion &             defocus &                zoom \\
\hline
Mask2Former+swinL &              $0.97$ &              $0.98$ &              $1.00$ &              $0.85$ &              $0.97$ &              $0.87$ &              $0.88$ &              $0.93$ &              $0.86$ &              $0.93$ &              $0.86$ &              $0.97$ &              $0.94$ &              $0.86$ &              $0.92$ &              $0.99$ &              $0.84$ \\
Mask2Former+R50   &              $0.96$ &              $0.98$ &              $1.00$ &              $0.85$ &              $0.93$ &              $0.81$ &              $0.79$ &              $0.92$ &              $0.84$ &              $0.90$ &              $0.80$ &              $0.93$ &              $0.88$ &              $0.79$ &              $0.86$ &              $0.98$ &              $0.85$ \\
\hline
ODISE+Caption     &     $\mathbf{0.99}$ &     $\mathbf{0.99}$ &     $\mathbf{1.00}$ &     $\mathbf{0.95}$ &     $\mathbf{0.99}$ &     $\mathbf{0.96}$ &     $\mathbf{0.95}$ &     $\mathbf{0.95}$ &     $\mathbf{0.95}$ &     $\mathbf{0.96}$ &     $\mathbf{0.96}$ &     $\mathbf{0.99}$ &     $\mathbf{0.99}$ &  $\underline{0.95}$ &  $\underline{0.98}$ &     $\mathbf{1.00}$ &     $\mathbf{0.95}$ \\
ODISE+Label       &  $\underline{0.99}$ &  $\underline{0.99}$ &  $\underline{1.00}$ &  $\underline{0.95}$ &  $\underline{0.99}$ &  $\underline{0.95}$ &  $\underline{0.95}$ &  $\underline{0.95}$ &  $\underline{0.95}$ &  $\underline{0.96}$ &  $\underline{0.95}$ &  $\underline{0.99}$ &  $\underline{0.98}$ &     $\mathbf{0.95}$ &     $\mathbf{0.98}$ &  $\underline{1.00}$ &  $\underline{0.94}$ \\
GroundedSam+swinB &              $0.98$ &              $0.99$ &              $1.00$ &              $0.92$ &              $0.98$ &              $0.93$ &              $0.93$ &              $0.94$ &              $0.92$ &              $0.95$ &              $0.93$ &              $0.99$ &              $0.97$ &              $0.93$ &              $0.96$ &              $0.99$ &              $0.91$ \\
\hline
\end{tabular}%
}
\end{table*}

\begin{table*}
\centering
    \caption{\textbf{Relative Robustness scores ($\gamma^r$) for instance segmentation} on models on the MS COCO-P and ADE20K-P. Models with the least relative drop in performance are in bold, and models that are second least are underlined.}
    \label{tab:relative_results}
    \resizebox{.99\linewidth}{!}{% 
    \begin{tabular}{l|cccc|cc|cc|cccc|cc|ccc}
\hline
 \multirow{2}{*}{\textbf{COCO} ($\gamma^r$)} & \multicolumn{4}{c|}{\textbf{Environment}} & \multicolumn{2}{c|}{\textbf{Digital}} & \multicolumn{2}{c|}{\textbf{Compression}} & \multicolumn{4}{c|}{\textbf{Pixel Noise}} & \multicolumn{2}{c|}{\textbf{Camera}} & \multicolumn{3}{c}{\textbf{Blur}} \\
 \cline{2-18}
 &            dark &          bright &                snow &                 fog &               shear &            contrast &                jpeg &            pixel. &             speckle &            gauss. &                shot &             impulse &              rotate &           translate &              motion &             defocus &                zoom \\
 \hline
Mask2Former+R50   &              $0.98$ &              $0.90$ &              $0.50$ &              $0.83$ &              $0.86$ &              $0.91$ &              $0.57$ &              $0.53$ &              $0.74$ &              $0.52$ &              $0.80$ &              $0.41$ &              $0.66$ &              $0.91$ &              $0.49$ &              $0.49$ &              $0.47$ \\
MaskDINO+R50      &              $0.98$ &              $0.90$ &              $0.49$ &              $0.84$ &              $0.86$ &              $0.92$ &              $0.56$ &              $0.51$ &              $0.73$ &              $0.52$ &              $0.79$ &              $0.45$ &              $0.65$ &              $0.90$ &              $0.50$ &              $0.52$ &              $0.47$ \\
Mask2Former+swinL &              $0.99$ &              $0.95$ &              $0.79$ &              $0.94$ &              $0.89$ &              $0.95$ &              $0.78$ &              $0.76$ &  $\underline{0.88}$ &              $0.73$ &  $\underline{0.90}$ &              $0.76$ &              $0.78$ &              $0.91$ &              $0.63$ &              $0.62$ &              $0.54$ \\
MaskDINO+swinL    &              $0.99$ &              $0.95$ &  $\underline{0.81}$ &              $0.94$ &              $0.88$ &              $0.95$ &              $0.79$ &  $\underline{0.78}$ &              $0.87$ &  $\underline{0.73}$ &              $0.90$ &  $\underline{0.76}$ &              $0.78$ &              $0.90$ &              $0.64$ &              $0.63$ &              $0.55$ \\
VitL-adapter      &              $0.99$ &              $0.95$ &              $0.73$ &              $0.92$ &              $0.87$ &  $\underline{0.96}$ &  $\underline{0.82}$ &              $0.78$ &              $0.87$ &              $0.68$ &              $0.89$ &              $0.70$ &              $0.76$ &              $0.90$ &              $0.64$ &              $0.62$ &              $0.55$ \\
\hline
ODISE+Caption     &     $\mathbf{1.00}$ &              $0.95$ &              $0.72$ &              $0.93$ &  $\underline{0.91}$ &              $0.94$ &              $0.64$ &              $0.67$ &              $0.87$ &              $0.69$ &              $0.89$ &              $0.67$ &     $\mathbf{0.79}$ &  $\underline{0.95}$ &              $0.63$ &              $0.62$ &              $0.55$ \\
ODISE+Label       &              $0.99$ &              $0.95$ &              $0.75$ &              $0.94$ &              $0.89$ &              $0.95$ &              $0.63$ &              $0.67$ &              $0.88$ &              $0.71$ &              $0.90$ &              $0.69$ &              $0.78$ &              $0.92$ &              $0.62$ &              $0.62$ &              $0.55$ \\
Prompt+SAM        &  $\underline{1.00}$ &  $\underline{0.95}$ &     $\mathbf{0.81}$ &     $\mathbf{0.96}$ &              $0.89$ &              $0.96$ &              $0.69$ &              $0.76$ &              $0.87$ &              $0.71$ &              $0.90$ &              $0.74$ &              $0.78$ &              $0.91$ &  $\underline{0.65}$ &     $\mathbf{0.65}$ &  $\underline{0.59}$ \\
InternImage-XL    &              $0.99$ &              $0.95$ &              $0.77$ &              $0.95$ &              $0.88$ &     $\mathbf{0.96}$ &     $\mathbf{0.82}$ &              $0.76$ &     $\mathbf{0.88}$ &     $\mathbf{0.75}$ &     $\mathbf{0.91}$ &     $\mathbf{0.78}$ &              $0.76$ &              $0.90$ &              $0.64$ &              $0.64$ &              $0.57$ \\
PAINTER           &              $0.98$ &              $0.94$ &              $0.65$ &              $0.89$ &     $\mathbf{0.93}$ &              $0.96$ &              $0.78$ &     $\mathbf{0.82}$ &              $0.86$ &              $0.69$ &              $0.89$ &              $0.69$ &  $\underline{0.78}$ &     $\mathbf{0.97}$ &     $\mathbf{0.72}$ &  $\underline{0.65}$ &     $\mathbf{0.60}$ \\
GroundedSam+swinB &              $0.99$ &     $\mathbf{0.96}$ &              $0.78$ &  $\underline{0.95}$ &              $0.90$ &              $0.96$ &              $0.71$ &              $0.74$ &              $0.86$ &              $0.69$ &              $0.88$ &              $0.71$ &              $0.76$ &              $0.92$ &              $0.63$ &              $0.63$ &              $0.58$ \\

\hline
\hline
 \multirow{2}{*}{\textbf{ADE20K} ($\gamma^r$)} & \multicolumn{4}{c|}{\textbf{Environment}} & \multicolumn{2}{c|}{\textbf{Digital}} & \multicolumn{2}{c|}{\textbf{Compression}} & \multicolumn{4}{c|}{\textbf{Pixel Noise}} & \multicolumn{2}{c|}{\textbf{Camera}} & \multicolumn{3}{c}{\textbf{Blur}} \\
 \cline{2-18}
 &            dark &          bright &                snow &                 fog &               shear &            contrast &                jpeg &            pixel. &             speckle &            gauss. &                shot &             impulse &              rotate &           translate &              motion &             defocus &                zoom \\
\hline
Mask2Former+swinL &  $\underline{0.92}$ &  $\underline{0.95}$ &  $\underline{1.00}$ &              $0.56$ &              $0.91$ &              $0.61$ &              $0.64$ &     $\mathbf{0.80}$ &              $0.59$ &     $\mathbf{0.79}$ &              $0.58$ &              $0.93$ &              $0.82$ &              $0.60$ &              $0.77$ &              $0.98$ &              $0.54$ \\
Mask2Former+R50   &              $0.84$ &              $0.91$ &              $0.99$ &              $0.44$ &              $0.74$ &              $0.27$ &              $0.20$ &  $\underline{0.68}$ &              $0.41$ &              $0.63$ &              $0.24$ &              $0.75$ &              $0.56$ &              $0.22$ &              $0.47$ &              $0.93$ &              $0.42$ \\
\hline
ODISE+Caption     &     $\mathbf{0.92}$ &     $\mathbf{0.95}$ &     $\mathbf{1.01}$ &     $\mathbf{0.65}$ &     $\mathbf{0.94}$ &  $\underline{0.68}$ &  $\underline{0.67}$ &              $0.67$ &     $\mathbf{0.68}$ &              $0.74$ &     $\mathbf{0.68}$ &     $\mathbf{0.95}$ &     $\mathbf{0.90}$ &  $\underline{0.64}$ &  $\underline{0.85}$ &     $\mathbf{0.99}$ &     $\mathbf{0.61}$ \\
ODISE+Label       &              $0.91$ &              $0.93$ &              $0.99$ &  $\underline{0.63}$ &  $\underline{0.92}$ &     $\mathbf{0.69}$ &     $\mathbf{0.68}$ &              $0.67$ &  $\underline{0.64}$ &  $\underline{0.74}$ &  $\underline{0.67}$ &              $0.92$ &  $\underline{0.88}$ &     $\mathbf{0.67}$ &     $\mathbf{0.86}$ &  $\underline{0.99}$ &  $\underline{0.59}$ \\
GroundedSam+swinB &              $0.87$ &              $0.93$ &              $0.98$ &              $0.55$ &              $0.91$ &              $0.57$ &              $0.60$ &              $0.67$ &              $0.57$ &              $0.70$ &              $0.59$ &  $\underline{0.93}$ &              $0.80$ &              $0.60$ &              $0.75$ &              $0.95$ &              $0.50$ \\
\hline
\end{tabular}%
}
\end{table*}

\begin{table*}
\centering
    \caption{\textbf{Relative robustness scores ($\gamma^r$) and Absolute robustness scores ($\gamma^a$) for each object super-category for instance segmentation on MS COCO-P.} Here the scores are averaged across all corruptions, severity for each model.}
    \label{tab:per_supercategory_coco_Relative_Robustness}
    \resizebox{\linewidth}{!}{\begin{tabular}{l|llllllllllll}
\hline
$\gamma^r$ &           accessory &              animal &           appliance &          electronic &                food &           furniture &              indoor &             kitchen &             outdoor &              person &              sports &             vehicle \\
\hline
Mask2Former+R50   &              $0.62$ &              $0.71$ &              $0.65$ &              $0.68$ &              $0.65$ &              $0.68$ &              $0.61$ &              $0.58$ &              $0.71$ &              $0.73$ &              $0.62$ &              $0.68$ \\
MaskDINO+R50      &              $0.63$ &              $0.71$ &              $0.65$ &              $0.68$ &              $0.64$ &              $0.68$ &              $0.61$ &              $0.58$ &              $0.73$ &              $0.74$ &              $0.64$ &              $0.68$ \\
Mask2Former+swinL &              $0.74$ &  $\underline{0.86}$ &              $0.80$ &              $0.80$ &              $0.82$ &              $0.84$ &              $0.76$ &  $\underline{0.72}$ &              $0.82$ &              $0.82$ &              $0.74$ &              $0.81$ \\
MaskDINO+swinL    &              $0.75$ &     $\mathbf{0.86}$ &              $0.83$ &     $\mathbf{0.81}$ &              $0.83$ &              $0.85$ &              $0.75$ &              $0.71$ &              $0.82$ &              $0.82$ &              $0.74$ &  $\underline{0.82}$ \\
VitL-adapter      &              $0.73$ &              $0.85$ &              $0.81$ &              $0.78$ &              $0.81$ &              $0.81$ &              $0.73$ &              $0.71$ &              $0.82$ &              $0.81$ &              $0.74$ &              $0.80$ \\
\hline
ODISE+Caption     &              $0.73$ &              $0.81$ &     $\mathbf{1.02}$ &              $0.75$ &              $0.81$ &              $0.81$ &              $0.75$ &              $0.70$ &              $0.80$ &              $0.82$ &              $0.71$ &              $0.80$ \\
ODISE+Label       &              $0.75$ &              $0.82$ &              $0.78$ &              $0.77$ &              $0.78$ &              $0.82$ &              $0.74$ &              $0.70$ &              $0.80$ &  $\underline{0.83}$ &              $0.71$ &              $0.81$ \\
Prompt+SAM        &  $\underline{0.76}$ &              $0.83$ &  $\underline{0.88}$ &  $\underline{0.81}$ &     $\mathbf{0.87}$ &     $\mathbf{0.92}$ &              $0.74$ &              $0.71$ &              $0.82$ &              $0.77$ &              $0.74$ &              $0.77$ \\
InternImage-XL    &              $0.75$ &              $0.85$ &              $0.86$ &              $0.80$ &              $0.82$ &              $0.84$ &              $0.75$ &     $\mathbf{0.72}$ &              $0.83$ &              $0.82$ &              $0.75$ &     $\mathbf{0.82}$ \\
PAINTER           &     $\mathbf{0.77}$ &              $0.83$ &              $0.76$ &              $0.76$ &              $0.78$ &              $0.78$ &     $\mathbf{0.80}$ &              $0.70$ &  $\underline{0.84}$ &     $\mathbf{0.87}$ &  $\underline{0.76}$ &              $0.81$ \\
GroundedSam+swinB &              $0.71$ &              $0.85$ &              $0.83$ &              $0.76$ &  $\underline{0.83}$ &  $\underline{0.89}$ &  $\underline{0.76}$ &              $0.70$ &     $\mathbf{1.10}$ &              $0.82$ &     $\mathbf{0.93}$ &              $0.78$ \\
\hline
\hline
$\gamma^a$ &           accessory &              animal &           appliance &          electronic &                food &           furniture &              indoor &             kitchen &             outdoor &              person &              sports &             vehicle \\
\hline
Mask2Former+R50   &              $0.87$ &              $0.83$ &              $0.83$ &              $0.83$ &              $0.87$ &              $0.89$ &              $0.88$ &              $0.87$ &              $0.87$ &              $0.87$ &              $0.86$ &              $0.86$ \\
MaskDINO+R50      &              $0.87$ &              $0.83$ &              $0.83$ &              $0.83$ &              $0.86$ &              $0.89$ &              $0.88$ &              $0.86$ &              $0.89$ &              $0.87$ &              $0.86$ &              $0.86$ \\
Mask2Former+swinL &              $0.89$ &              $0.92$ &              $0.88$ &              $0.89$ &              $0.92$ &              $0.94$ &              $0.91$ &              $0.89$ &              $0.92$ &              $0.90$ &              $0.89$ &              $0.91$ \\
MaskDINO+swinL    &              $0.89$ &              $0.92$ &              $0.90$ &  $\underline{0.89}$ &              $0.92$ &              $0.94$ &              $0.90$ &              $0.88$ &              $0.91$ &              $0.90$ &              $0.89$ &              $0.91$ \\
VitL-adapter      &              $0.90$ &              $0.91$ &              $0.90$ &              $0.88$ &              $0.93$ &              $0.93$ &              $0.90$ &              $0.89$ &              $0.92$ &              $0.90$ &              $0.90$ &              $0.91$ \\
\hline
ODISE+Caption     &  $\underline{0.92}$ &              $0.89$ &  $\underline{0.93}$ &              $0.89$ &              $0.94$ &              $0.95$ &  $\underline{0.93}$ &  $\underline{0.93}$ &              $0.93$ &              $0.91$ &              $0.90$ &  $\underline{0.92}$ \\
ODISE+Label       &              $0.91$ &              $0.89$ &              $0.89$ &              $0.88$ &              $0.92$ &              $0.93$ &              $0.91$ &              $0.90$ &              $0.91$ &              $0.91$ &              $0.89$ &              $0.91$ \\
Prompt+SAM        &              $0.90$ &              $0.90$ &     $\mathbf{0.94}$ &              $0.89$ &  $\underline{0.95}$ &     $\mathbf{0.98}$ &              $0.91$ &              $0.90$ &              $0.92$ &              $0.89$ &  $\underline{0.90}$ &              $0.90$ \\
InternImage-XL    &              $0.90$ &              $0.91$ &              $0.92$ &              $0.89$ &              $0.92$ &              $0.94$ &              $0.91$ &              $0.89$ &              $0.92$ &              $0.90$ &              $0.90$ &              $0.92$ \\
PAINTER           &     $\mathbf{0.96}$ &     $\mathbf{0.93}$ &              $0.88$ &     $\mathbf{0.92}$ &     $\mathbf{0.96}$ &              $0.95$ &     $\mathbf{0.97}$ &     $\mathbf{0.96}$ &     $\mathbf{0.96}$ &     $\mathbf{0.96}$ &     $\mathbf{0.95}$ &     $\mathbf{0.94}$ \\
GroundedSam+swinB &              $0.91$ &  $\underline{0.92}$ &              $0.91$ &              $0.88$ &              $0.94$ &  $\underline{0.97}$ &              $0.92$ &              $0.91$ &  $\underline{0.95}$ &  $\underline{0.92}$ &              $0.79$ &              $0.91$ \\

\hline
\end{tabular}}
\end{table*}

\subsection{Semantic Segmentation}
\begin{figure*}
    \centering
    \includegraphics[width=\linewidth]{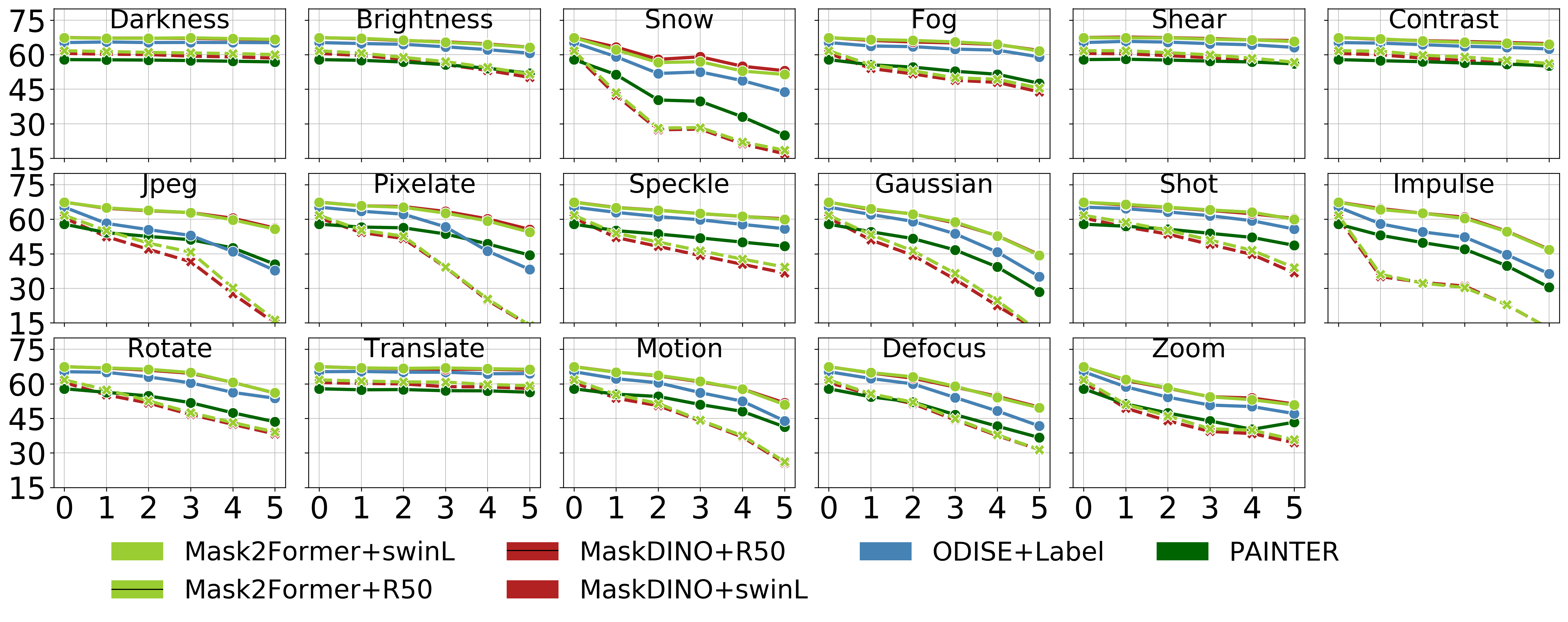}
    \caption{\textbf{Results for each corruption and each severity for semantic segmentation measured on the MS COCO-P dataset}. x-axis: Severity ranges from 0 (no corruption) to 5 (most corruption). y-axis: model performance measured by mean intersection-over-union (mIoU).}
    \label{fig:coco_semantic_line_plots}
\end{figure*}

\begin{figure*}[t!]
    \centering
    \includegraphics[width=\linewidth]{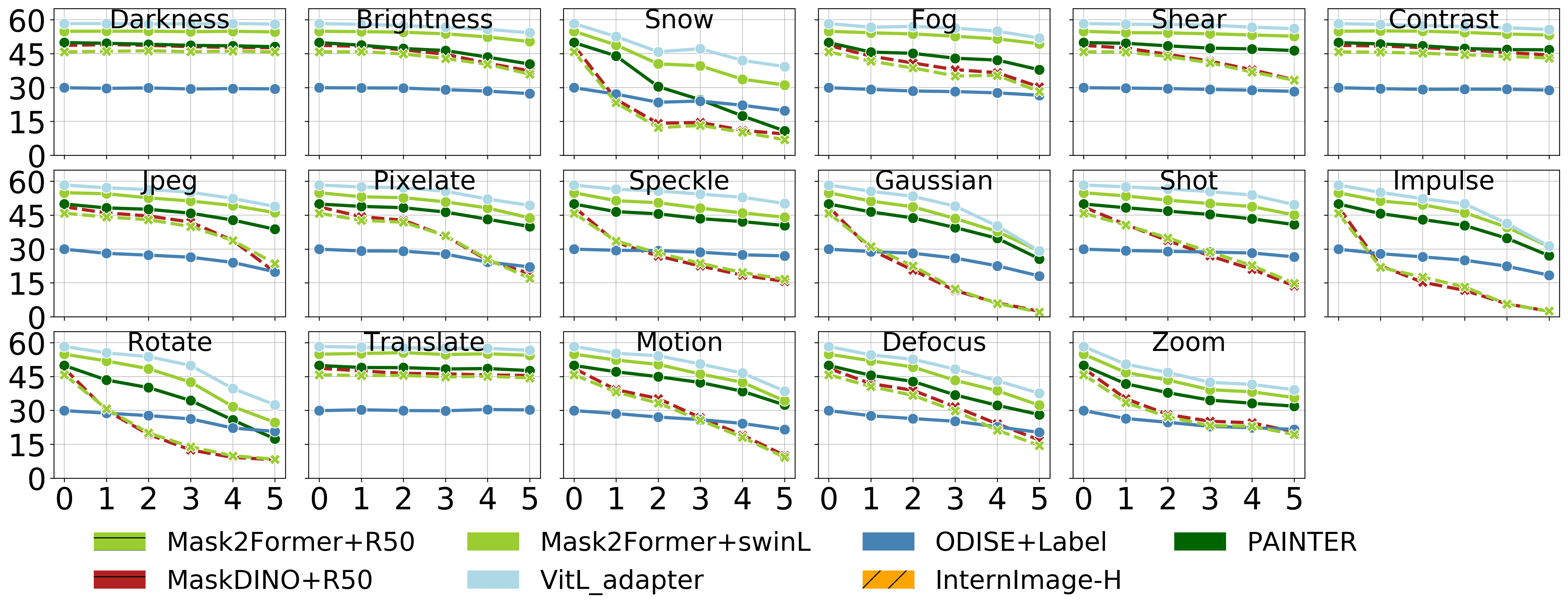}
    \caption{\textbf{Results for each corruption and each severity for semantic segmentation measured on the ADE20K-P dataset}. x-axis: Severity ranges from 0 (no corruption) to 5 (most corruption). y-axis: model performance measured by mean intersection-over-union (mIoU).}
    \label{fig:ade20k_semantic_line_plots}
\end{figure*}

Table \ref{tab:semantic_relative_scores} and \ref{tab:semantic_absolute_scores} show relative robustness ($\gamma^r$) and absolute robustness ($\gamma^a$) scores for the selected models on semantic segmentation. While comparing, we do find that all selected models are typically more robust to semantic segmentation as opposed to instance segmentation, but still multimodal models perform poorly for compression, snow and certain noises in comparison to transformer based unimodal in MS COCO-P dataset. We observe that CNN models on the ADE20K-P dataset are even less robust compared to MS COCO-P. Additionally, the ODISE model is more relatively robust on ADE20K-P, where it is evaluated zero-shot, as compared to MS COCO-P. To better observe performance across varying severity for each corruption, we visualize results for COCO-P in \ref{fig:coco_semantic_line_plots} and ADE20K-P in \ref{fig:ade20k_semantic_line_plots}. For both datasets, but especially ADE20K-P, we see CNN-based backbones for unimodal models have a much steeper decline as corruption severity increases, most noticeably for noise-based corruptions. This may indicate CNN-based architectures are more sensitive to noise-based corruptions.

\begin{table*}
% %\vspace{-10pt}
\centering
    \caption{\textbf{Relative Robustness scores ($\gamma^r$) for models on the MS COCO-P and ADE20K-P dataset for semantic segmentation}. Models with the least relative drop in performance are in bold, and models that are second least are underlined.}
    \label{tab:semantic_relative_scores}
    \resizebox{\linewidth}{!}{\begin{tabular}{l|cccc|cc|cc|cccc|cc|ccc}
\hline
 \multirow{2}{*}{\textbf{COCO} ($\gamma^r$)} & \multicolumn{4}{c|}{\textbf{Environment}} & \multicolumn{2}{c|}{\textbf{Digital}} & \multicolumn{2}{c|}{\textbf{Compression}} & \multicolumn{4}{c|}{\textbf{Pixel Noise}} & \multicolumn{2}{c|}{\textbf{Camera}} & \multicolumn{3}{c}{\textbf{Blur}} \\
 \cline{2-18}
 &            dark &          bright &                snow &                 fog &               shear &            contrast &                jpeg &            pixel. &             speckle &            gauss. &                shot &             impulse &              rotate &           translate &              motion &             defocus &                zoom \\
 \hline
Mask2Former+R50   &              $0.98$ &              $0.92$ &              $0.46$ &              $0.82$ &              $0.97$ &              $0.95$ &              $0.64$ &              $0.60$ &              $0.75$ &              $0.56$ &              $0.81$ &              $0.43$ &              $0.78$ &              $0.98$ &              $0.70$ &              $0.72$ &              $0.69$ \\
MaskDINO+R50      &              $0.98$ &              $0.91$ &              $0.45$ &              $0.81$ &              $0.96$ &              $0.95$ &              $0.61$ &              $0.61$ &              $0.73$ &              $0.54$ &              $0.79$ &              $0.44$ &              $0.77$ &              $0.98$ &              $0.70$ &              $0.73$ &              $0.68$ \\
Mask2Former+swinL &  $\underline{1.00}$ &     $\mathbf{0.97}$ &  $\underline{0.83}$ &     $\mathbf{0.96}$ &              $0.99$ &              $0.97$ &  $\underline{0.91}$ &  $\underline{0.91}$ &  $\underline{0.93}$ &     $\mathbf{0.84}$ &     $\mathbf{0.95}$ &  $\underline{0.86}$ &     $\mathbf{0.93}$ &  $\underline{0.99}$ &  $\underline{0.89}$ &     $\mathbf{0.86}$ &  $\underline{0.83}$ \\
MaskDINO+swinL    &              $0.99$ &  $\underline{0.97}$ &     $\mathbf{0.86}$ &  $\underline{0.96}$ &     $\mathbf{0.99}$ &  $\underline{0.97}$ &     $\mathbf{0.91}$ &     $\mathbf{0.92}$ &     $\mathbf{0.93}$ &  $\underline{0.84}$ &  $\underline{0.94}$ &     $\mathbf{0.86}$ &  $\underline{0.93}$ &              $0.98$ &     $\mathbf{0.89}$ &  $\underline{0.86}$ &     $\mathbf{0.83}$ \\
\hline
ODISE+Caption     &              $0.99$ &              $0.96$ &              $0.76$ &              $0.93$ &              $0.98$ &              $0.97$ &              $0.77$ &              $0.82$ &              $0.90$ &              $0.76$ &              $0.92$ &              $0.73$ &              $0.89$ &              $0.99$ &              $0.83$ &              $0.80$ &              $0.79$ \\
ODISE+Label       &     $\mathbf{1.00}$ &              $0.97$ &              $0.78$ &              $0.95$ &  $\underline{0.99}$ &     $\mathbf{0.98}$ &              $0.77$ &              $0.82$ &              $0.91$ &              $0.78$ &              $0.93$ &              $0.75$ &              $0.91$ &     $\mathbf{0.99}$ &              $0.84$ &              $0.82$ &              $0.80$ \\
PAINTER           &              $0.99$ &              $0.95$ &              $0.65$ &              $0.91$ &              $0.99$ &              $0.97$ &              $0.85$ &              $0.90$ &              $0.89$ &              $0.76$ &              $0.92$ &              $0.76$ &              $0.88$ &              $0.99$ &              $0.87$ &              $0.80$ &              $0.78$ \\

\hline
\hline
 \multirow{2}{*}{\textbf{ADE20K} ($\gamma^r$)} & \multicolumn{4}{c|}{\textbf{Environment}} & \multicolumn{2}{c|}{\textbf{Digital}} & \multicolumn{2}{c|}{\textbf{Compression}} & \multicolumn{4}{c|}{\textbf{Pixel Noise}} & \multicolumn{2}{c|}{\textbf{Camera}} & \multicolumn{3}{c}{\textbf{Blur}} \\
 \cline{2-18}
 &            dark &          bright &                snow &                 fog &               shear &            contrast &                jpeg &            pixel. &             speckle &            gauss. &                shot &             impulse &              rotate &           translate &              motion &             defocus &                zoom \\
\hline
Mask2Former+R50   &     $\mathbf{1.00}$ &              $0.92$ &              $0.29$ &              $0.78$ &              $0.88$ &              $0.97$ &              $0.80$ &              $0.71$ &              $0.53$ &              $0.32$ &              $0.62$ &              $0.26$ &              $0.36$ &              $0.98$ &              $0.54$ &              $0.62$ &              $0.55$ \\
MaskDINO+R50      &              $0.99$ &              $0.90$ &              $0.30$ &              $0.78$ &              $0.84$ &              $0.96$ &              $0.76$ &              $0.69$ &              $0.48$ &              $0.29$ &              $0.56$ &              $0.24$ &              $0.33$ &              $0.95$ &              $0.54$ &              $0.63$ &              $0.55$ \\
Mask2Former+swinL &  $\underline{1.00}$ &     $\mathbf{0.97}$ &              $0.71$ &     $\mathbf{0.95}$ &  $\underline{0.98}$ &     $\mathbf{0.99}$ &  $\underline{0.92}$ &              $0.91$ &              $0.87$ &              $0.77$ &              $0.91$ &              $0.79$ &              $0.73$ &  $\underline{1.00}$ &              $0.82$ &              $0.79$ &              $0.74$ \\
VitL-adapter      &              $1.00$ &  $\underline{0.97}$ &  $\underline{0.78}$ &  $\underline{0.95}$ &     $\mathbf{0.98}$ &              $0.98$ &     $\mathbf{0.92}$ &     $\mathbf{0.93}$ &  $\underline{0.92}$ &              $0.78$ &  $\underline{0.94}$ &              $0.79$ &              $0.79$ &              $0.99$ &  $\underline{0.84}$ &  $\underline{0.81}$ &              $0.76$ \\
\hline
ODISE+Label       &              $0.99$ &              $0.97$ &     $\mathbf{0.78}$ &              $0.94$ &              $0.97$ &              $0.98$ &              $0.84$ &              $0.88$ &     $\mathbf{0.95}$ &     $\mathbf{0.83}$ &     $\mathbf{0.95}$ &  $\underline{0.80}$ &     $\mathbf{0.84}$ &     $\mathbf{1.01}$ &     $\mathbf{0.85}$ &     $\mathbf{0.82}$ &     $\mathbf{0.79}$ \\
InternImage-H     &              $1.00$ &              $0.96$ &              $0.73$ &              $0.94$ &              $0.97$ &  $\underline{0.98}$ &              $0.90$ &  $\underline{0.91}$ &              $0.89$ &  $\underline{0.79}$ &              $0.92$ &     $\mathbf{0.81}$ &  $\underline{0.80}$ &              $0.98$ &              $0.82$ &              $0.76$ &  $\underline{0.78}$ \\
PAINTER           &              $0.98$ &              $0.91$ &              $0.51$ &              $0.86$ &              $0.96$ &              $0.96$ &              $0.89$ &              $0.91$ &              $0.87$ &              $0.76$ &              $0.90$ &              $0.77$ &              $0.65$ &              $0.97$ &              $0.82$ &              $0.74$ &              $0.72$ \\
\hline
\end{tabular}}
\end{table*}

\begin{table*}
\centering
    \caption{\textbf{Absolute Robustness scores ($\gamma^a$) for models on the MS COCO-P and ADE20k-P dataset for semantic segmentation}. Models with the least relative drop in performance are in bold, and models that are second least are underlined.}
    \label{tab:semantic_absolute_scores}
    \resizebox{\linewidth}{!}{\begin{tabular}{l|cccc|cc|cc|cccc|cc|ccc}
\hline
 \multirow{2}{*}{\textbf{COCO} ($\gamma^a$)} & \multicolumn{4}{c|}{\textbf{Environment}} & \multicolumn{2}{c|}{\textbf{Digital}} & \multicolumn{2}{c|}{\textbf{Compression}} & \multicolumn{4}{c|}{\textbf{Pixel Noise}} & \multicolumn{2}{c|}{\textbf{Camera}} & \multicolumn{3}{c}{\textbf{Blur}} \\
 \cline{2-18}
 &            dark &          bright &                snow &                 fog &               shear &            contrast &                jpeg &            pixel. &             speckle &            gauss. &                shot &             impulse &              rotate &           translate &              motion &             defocus &                zoom \\
 \hline
Mask2Former+R50   &              $0.99$ &              $0.95$ &              $0.66$ &              $0.89$ &              $0.98$ &              $0.97$ &              $0.78$ &              $0.76$ &              $0.85$ &              $0.73$ &              $0.88$ &              $0.65$ &              $0.86$ &              $0.99$ &              $0.81$ &              $0.83$ &              $0.81$ \\
MaskDINO+R50      &              $0.99$ &              $0.95$ &              $0.67$ &              $0.89$ &              $0.98$ &              $0.97$ &              $0.76$ &              $0.76$ &              $0.84$ &              $0.72$ &              $0.88$ &              $0.66$ &              $0.86$ &              $0.99$ &              $0.82$ &              $0.83$ &              $0.81$ \\
Mask2Former+swinL &  $\underline{1.00}$ &     $\mathbf{0.98}$ &  $\underline{0.89}$ &     $\mathbf{0.98}$ &              $0.99$ &              $0.98$ &  $\underline{0.94}$ &              $0.94$ &  $\underline{0.95}$ &     $\mathbf{0.89}$ &     $\mathbf{0.96}$ &  $\underline{0.90}$ &     $\mathbf{0.96}$ &              $0.99$ &  $\underline{0.92}$ &     $\mathbf{0.91}$ &              $0.88$ \\
MaskDINO+swinL    &              $1.00$ &  $\underline{0.98}$ &     $\mathbf{0.90}$ &  $\underline{0.97}$ &     $\mathbf{0.99}$ &              $0.98$ &     $\mathbf{0.94}$ &     $\mathbf{0.95}$ &     $\mathbf{0.95}$ &  $\underline{0.89}$ &  $\underline{0.96}$ &     $\mathbf{0.91}$ &  $\underline{0.95}$ &              $0.99$ &     $\mathbf{0.92}$ &  $\underline{0.91}$ &  $\underline{0.88}$\\ 
\hline
ODISE+Caption     &              $1.00$ &              $0.98$ &              $0.87$ &              $0.96$ &              $0.99$ &              $0.98$ &              $0.88$ &              $0.90$ &              $0.95$ &              $0.88$ &              $0.96$ &              $0.86$ &              $0.94$ &  $\underline{0.99}$ &              $0.91$ &              $0.90$ &     $\mathbf{0.89}$ \\
ODISE+Label       &     $\mathbf{1.00}$ &              $0.98$ &              $0.86$ &              $0.97$ &  $\underline{0.99}$ &  $\underline{0.98}$ &              $0.85$ &              $0.88$ &              $0.94$ &              $0.86$ &              $0.96$ &              $0.84$ &              $0.94$ &     $\mathbf{1.00}$ &              $0.90$ &              $0.88$ &              $0.87$ \\
PAINTER           &              $1.00$ &              $0.97$ &              $0.80$ &              $0.95$ &              $0.99$ &     $\mathbf{0.99}$ &              $0.91$ &  $\underline{0.94}$ &              $0.94$ &              $0.86$ &              $0.96$ &              $0.86$ &              $0.93$ &              $0.99$ &              $0.92$ &              $0.88$ &              $0.87$ \\

\hline
\hline
 \multirow{2}{*}{\textbf{ADE20K} ($\gamma^a$)} & \multicolumn{4}{c|}{\textbf{Environment}} & \multicolumn{2}{c|}{\textbf{Digital}} & \multicolumn{2}{c|}{\textbf{Compression}} & \multicolumn{4}{c|}{\textbf{Pixel Noise}} & \multicolumn{2}{c|}{\textbf{Camera}} & \multicolumn{3}{c}{\textbf{Blur}} \\
 \cline{2-18}
 &            dark &          bright &                snow &                 fog &               shear &            contrast &                jpeg &            pixel. &             speckle &            gauss. &                shot &             impulse &              rotate &           translate &              motion &             defocus &                zoom \\
\hline
Mask2Former+R50   &     $\mathbf{1.00}$ &              $0.96$ &              $0.67$ &              $0.90$ &              $0.94$ &              $0.99$ &              $0.91$ &              $0.87$ &              $0.79$ &              $0.69$ &              $0.83$ &              $0.66$ &              $0.71$ &              $0.99$ &              $0.79$ &              $0.83$ &              $0.79$ \\
MaskDINO+R50      &              $1.00$ &              $0.95$ &              $0.66$ &              $0.89$ &              $0.92$ &              $0.98$ &              $0.89$ &              $0.85$ &              $0.75$ &              $0.65$ &              $0.79$ &              $0.63$ &              $0.67$ &              $0.98$ &              $0.77$ &              $0.82$ &              $0.78$ \\
Mask2Former+swinL &              $1.00$ &              $0.98$ &              $0.84$ &              $0.97$ &              $0.99$ &  $\underline{0.99}$ &     $\mathbf{0.96}$ &              $0.95$ &              $0.93$ &              $0.87$ &              $0.95$ &              $0.89$ &              $0.85$ &              $1.00$ &              $0.90$ &              $0.88$ &              $0.86$ \\
VitL-adapter      &              $1.00$ &              $0.98$ &              $0.87$ &              $0.97$ &              $0.99$ &              $0.99$ &  $\underline{0.96}$ &  $\underline{0.96}$ &              $0.96$ &              $0.87$ &              $0.96$ &              $0.88$ &              $0.88$ &              $0.99$ &              $0.91$ &              $0.89$ &              $0.86$ \\
\hline
ODISE+Caption     &  $\underline{1.00}$ &  $\underline{0.99}$ &  $\underline{0.91}$ &  $\underline{0.98}$ &  $\underline{0.99}$ &     $\mathbf{1.00}$ &              $0.95$ &              $0.96$ &  $\underline{0.97}$ &  $\underline{0.94}$ &  $\underline{0.98}$ &  $\underline{0.94}$ &  $\underline{0.94}$ &     $\mathbf{1.00}$ &     $\mathbf{0.96}$ &  $\underline{0.94}$ &     $\mathbf{0.94}$ \\
ODISE+Label       &              $1.00$ &     $\mathbf{0.99}$ &     $\mathbf{0.93}$ &     $\mathbf{0.98}$ &     $\mathbf{0.99}$ &              $0.99$ &              $0.95$ &     $\mathbf{0.97}$ &     $\mathbf{0.98}$ &     $\mathbf{0.95}$ &     $\mathbf{0.98}$ &     $\mathbf{0.94}$ &     $\mathbf{0.95}$ &  $\underline{1.00}$ &  $\underline{0.96}$ &     $\mathbf{0.95}$ &  $\underline{0.94}$ \\
InternImage-H     &              $1.00$ &              $0.98$ &              $0.84$ &              $0.96$ &              $0.98$ &              $0.99$ &              $0.94$ &              $0.95$ &              $0.94$ &              $0.87$ &              $0.95$ &              $0.89$ &              $0.88$ &              $0.99$ &              $0.89$ &              $0.86$ &              $0.87$ \\
PAINTER           &              $0.99$ &              $0.95$ &              $0.76$ &              $0.93$ &              $0.98$ &              $0.98$ &              $0.95$ &              $0.95$ &              $0.94$ &              $0.88$ &              $0.95$ &              $0.88$ &              $0.82$ &              $0.99$ &              $0.91$ &              $0.87$ &              $0.86$ \\

\hline
\end{tabular}}
% \{-10pt}
\end{table*}

\begin{figure*}[t!]
    \centering
    \includegraphics[width=\linewidth]{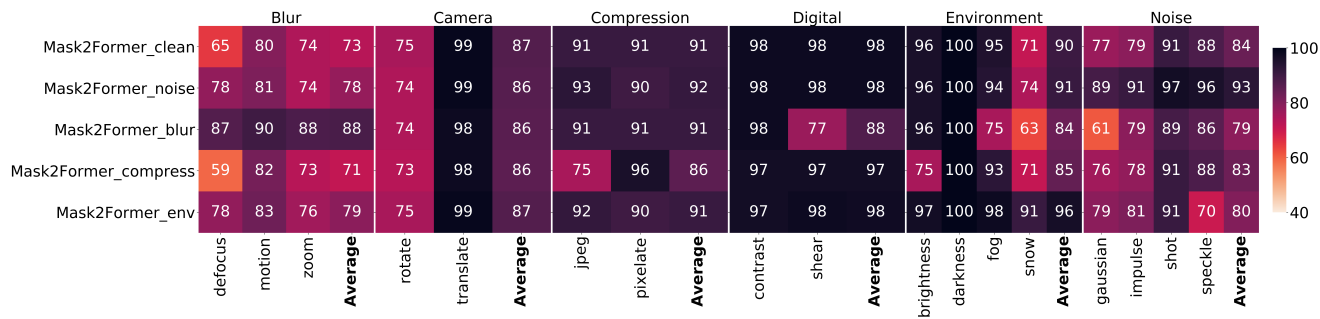}
    \caption{\textbf{Fine-tuned performance of Mask2Former on semantic segmentation for the augmented ADE20K-P dataset}. Here y-axis denotes augmentation used for fine-tuning (expect first row) and x-axis denotes models' relative robustness $\gamma^r$ for each corruption averaged over severity.}
    \label{fig:mask2former_finetune}
\end{figure*}

\begin{figure*}[t!]
    \centering
    \includegraphics[width=\linewidth]{vit_finetune_heatmap.png}
    \caption{\textbf{Fine-tuned performance of ViT-Adapter on semantic segmentation for the ADE20K-P dataset on}. Y-axis: Augmentation used for fine-tuning (expect first row). X-axis: model Relative Robustness score for each corruption averaged over severity.}
    \label{fig:vit_adapter_finetune}
\end{figure*}

\subsection{Fine-tuning on Corrupted Dataset}
The fine-tuning dataset comprises a subset of the ADE20K training dataset, consisting of 8000 images, which is consistent for all fine-tuning. The first 2000 are clean, while the remaining 6000 are randomly augmented using perturbations from the specific category we are targeting. Figure respectively shows the performance of Mask2Former and ViT-Adapter.

\subsection{Qualitative Examples}
\begin{figure*}
    \centering
    \includegraphics[width=\linewidth]{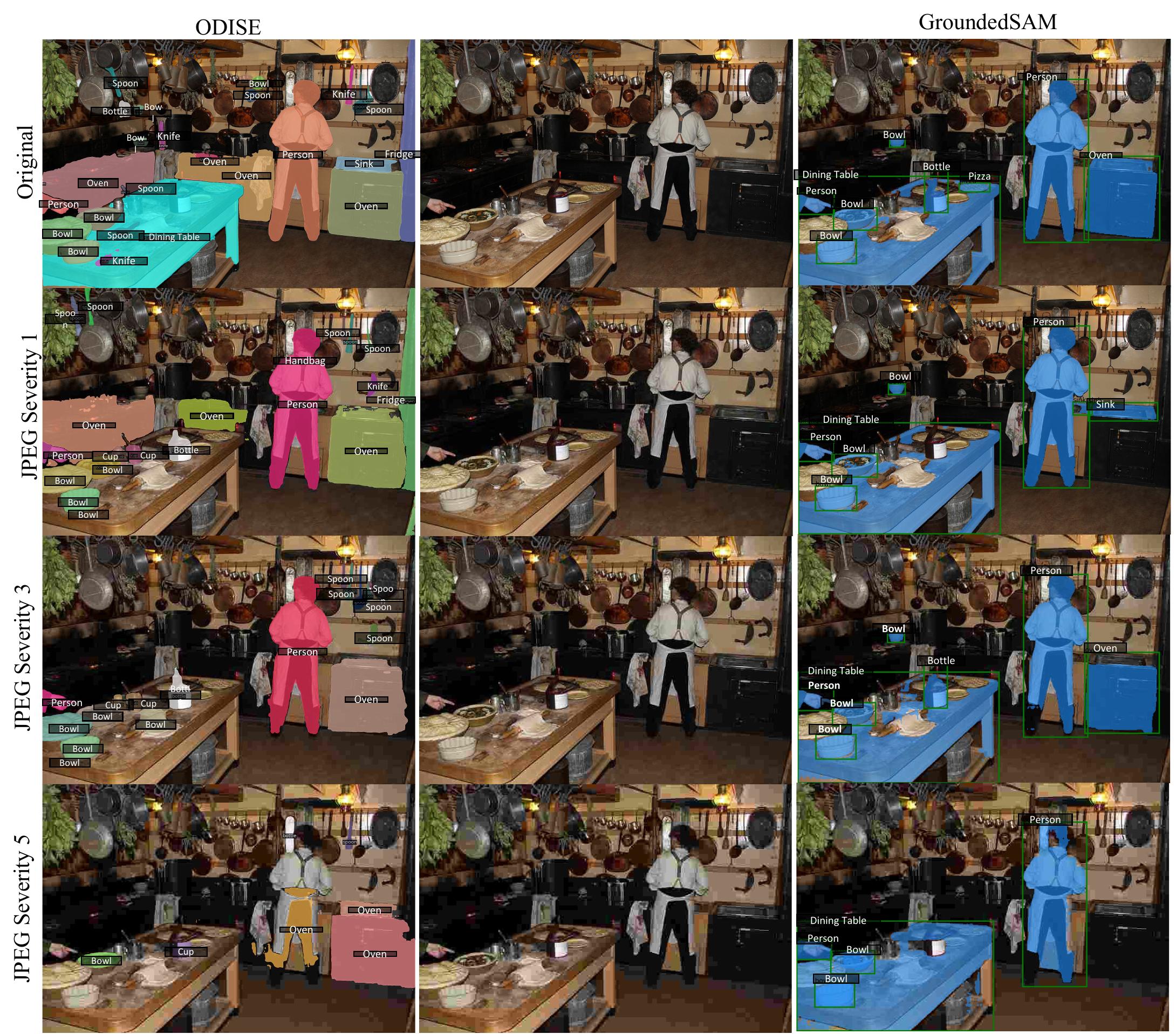}
    \caption{\textbf{Visual example from the \textbf{COCO-P} dataset for \textbf{JPEG} compression under varying levels of severity.} The left shows results for ODISE, middle shows the original images, and the right shows GroundedSAM. We again see as severity increases, both models mask quality decreases but ODISE additionally misclassifies objects, such as ``person'' to ``oven''.}
    \label{fig:jpeg_vfm_examples}
\end{figure*}

\begin{figure*}
    \centering
    \includegraphics[width=\linewidth]{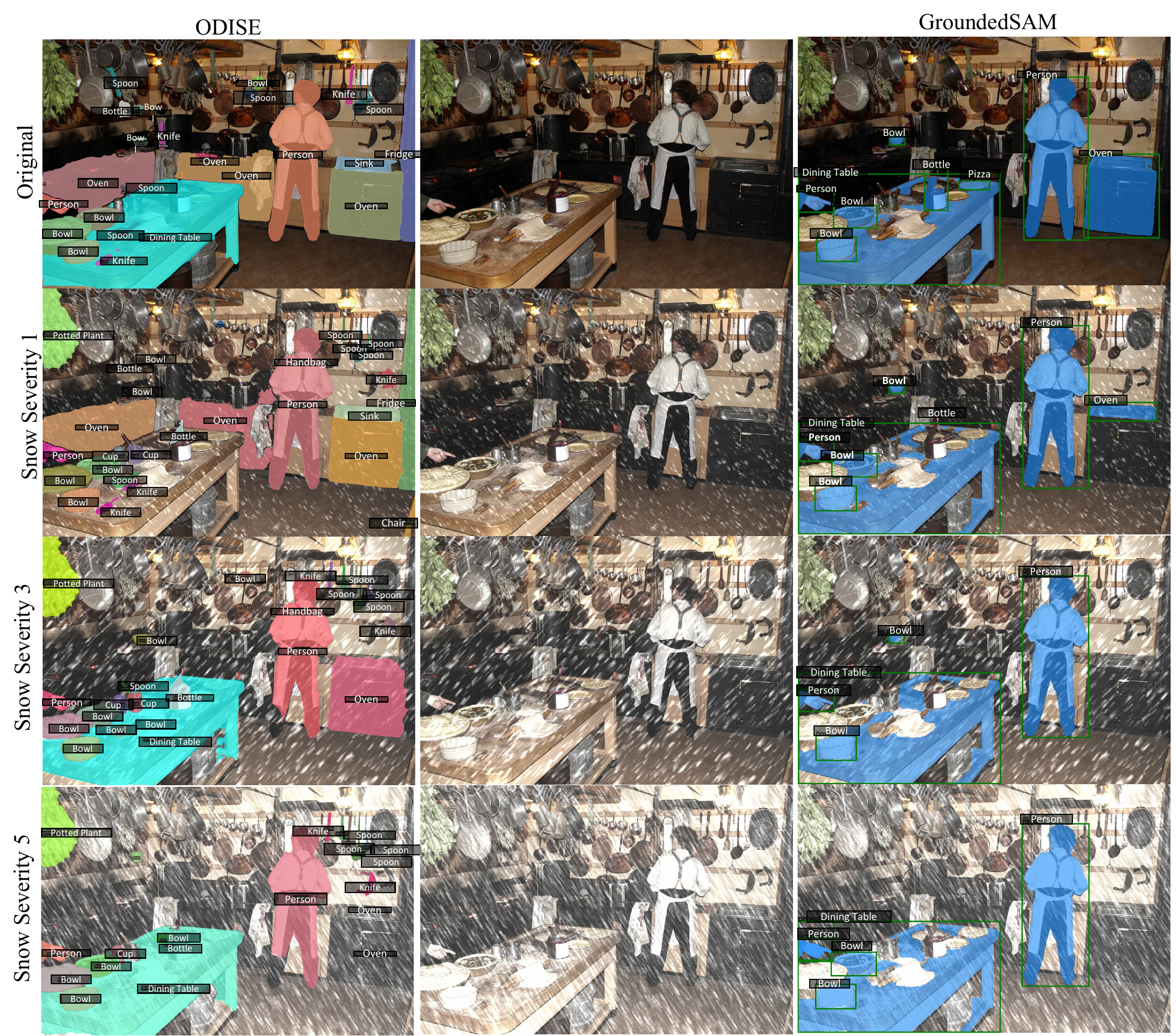}
    \caption{\textbf{Visual example from the \textbf{MS COCO-P} dataset for \textbf{Snow} corruption under varying levels of severity}. The left shows results for ODISE, middle shows the original images, and the right shows GroundedSAM. We again see as severity increases, both models mask quality decreases but ODISE additionally incorrectly classifies objects, such as ``person'' to ``oven''.}
    \label{fig:snow_vfm_examples}
\end{figure*}

\begin{figure*}
    \centering
    \includegraphics[width=\linewidth]{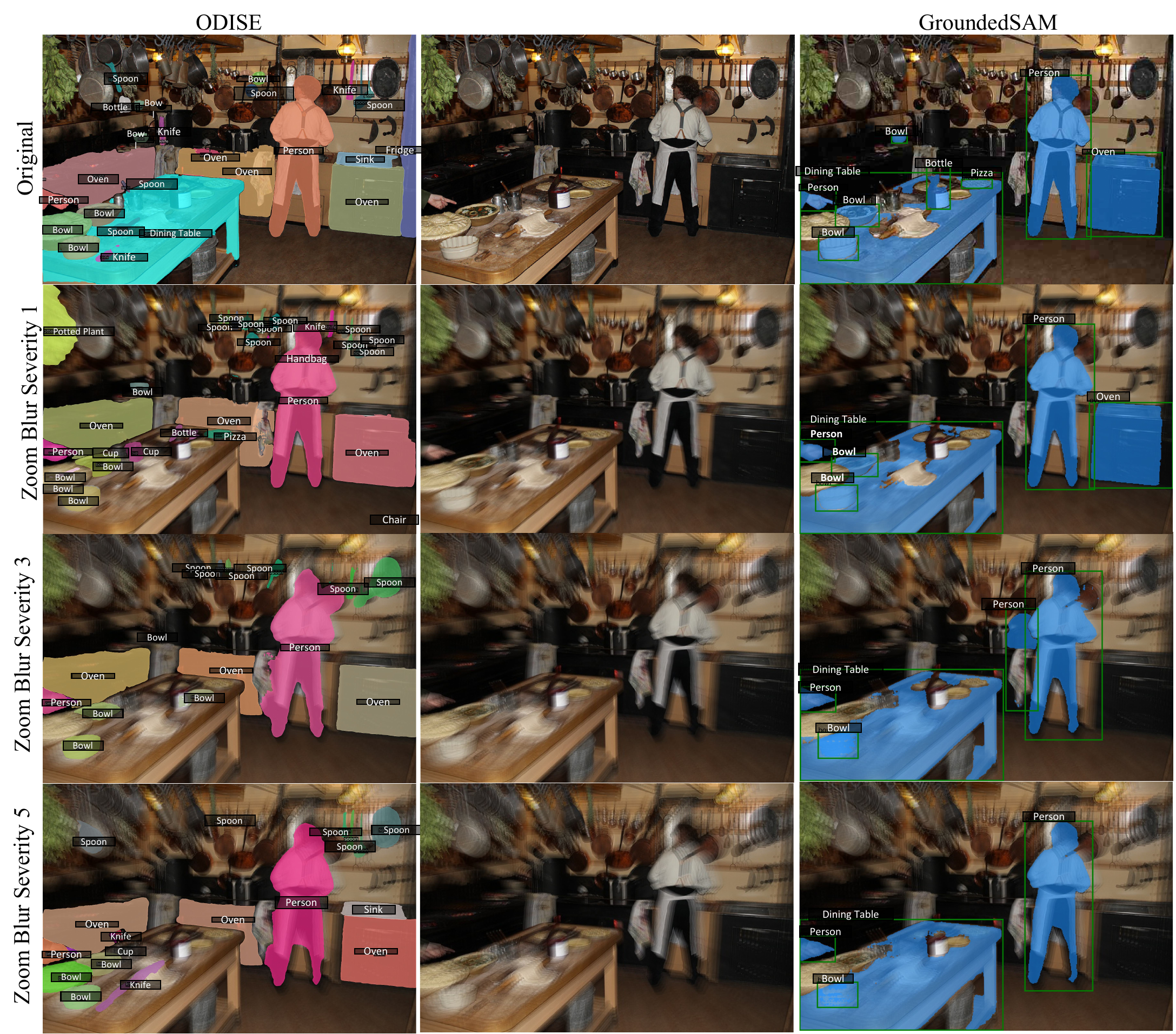}
    \caption{\textbf{Visual example from the \textbf{COCO-P} dataset for \textbf{Zoom Blur} corruption under varying levels of severity.} The left shows results for ODISE, middle shows the original images, and the right shows GroundedSAM. We again see as severity increases, both models mask quality decreases but ODISE additionally incorrectly classifies objects, such as ``person'' to ``oven''.}
    \label{fig:zoom_vfm_examples}
\end{figure*}

We show examples of model predictions in Figures \ref{fig:jpeg_vfm_examples}, \ref{fig:zoom_vfm_examples} and \ref{fig:snow_vfm_examples}. Figure \ref{fig:jpeg_vfm_examples} shows an image from the COCO-P dataset under \textit{JPEG} compression with severity 1, 3, and 5. As severity increases, mask quality and the number of objects decreases. This is more noticeable with ODISE where it additionally classifies objects. Figure \ref{fig:snow_vfm_examples} shows the same but under the \textit{snow} corruption. Models are typically more robust to \textit{snow} as compared to \textit{JPEG}, but show some decrease in performance as severity increases as shown in Figure \ref{fig:coco_corruption_line_plots}. Here we see mask quality persist but the number of smaller objects classified and masked decrease. Figure \ref{fig:zoom_vfm_examples} shows the same but for \textit{zoom blur}, a corruption all models are low in robustness to. Again we see as severity increases, ODISE misclassifies some objects. However, even with the low robustness to blur, we see the mask quality is still visually higher when compared to \textit{JPEG}.

\end{document}